# An Empirical Analysis of Deep Learning Methods for Small Object Detection from Satellite Imagery

Xiaohui Yuan[a], Aniv Chakravarty[a], Elinor M. Lichtenberg[b], Lichuan Gu[c], Zhenchun Wei[d], Tian Chen[d]

[a]*Department of Computer Science and Engineering, University of North Texas, Denton, 76207, Texas, USA*
[b]*Department of Biological Sciences and Advanced Environmental Research Institute, University of North Texas, Denton, 76203, Texas, USA*
[c]*School of Artificial Intelligence and Computer Science, Anhui Agricultural University, Hefei, 230009, Anhui, China*
[d]*School of Computer Science and Information, Hefei University of Technology, Hefei, 230009, Anhui, China*

**Abstract**

Despite a substantial body of literature on object detection, there is a notable lack of empirical studies on detecting small objects. Additionally, the definition of a small object remains unclear. This paper presents a thorough evaluation of six state-of-the-art deep learning methods for small object detection from satellite imagery. Three public high-resolution datasets are used to understand various influential aspects and the generalization ability. Among the six methods, YOLOv11 achieves a balanced performance for localization and adaptability, while Faster R-CNN maintains consistent detection coverage. Anchor box-based methods require extensive fine-tuning, whereas transformer-based methods demand greater computational resources to achieve competitive results. In addition, anchor-based methods, including SSD, Faster R-CNN, and Cascade R-CNN, are sensitive to the anchor box size, and, for small object detection, a small to moderate size is preferred. Both deformable and RT-DETR methods are susceptible to overfitting. RT-DETR exhibits superior detection in partial occlusion scenarios, particularly through vegetation and shadows, whereas deformable DETR struggles to identify individual small objects in dense clusters. Comparing computational efficiency with a batch size of one reveals that RT-DETR and YOLOv11 are more training-intensive, with optimizations focused on inference. Methods such as Faster R-CNN have a larger memory footprint but lower computational costs and time requirements.

*Keywords:* Object detection, Machine learning, Remote sensing

## 1. Introduction

Object detection provides cues for many computer vision applications, such as object recognition and tracking movements of targets in videos [1, 2, 3]. There are many learning-based methods developed in recent years, such as You Only Look Once (YOLO) [4], Single Shot MultiBox Detection (SSD) [5], and derivations of Region-Based Convolutional Neural Networks (R-CNN) [6] (e.g., Faster R-CNN [7] and Cascade R-CNN [8]). Recent developments with transformers, e.g., detection transformer (DETR) [9], improved object detection accuracy but with a high computational overhead. However, object detection methods for satellite imagery analysis face unique challenges. Objects in satellite imagery usually have small footprints with a diverse distribution and are distinct from terrain and morphology at different geographical locations.

Small objects are usually defined by their absolute size (number of pixels) or by their relative size in an image [10, 11]. Detecting the small objects is difficult because they lack discriminative features, which leads to several problems: missing critical features, confusion with parts of larger objects, and weak contributions to the learning process. Texture, region, and shape cues that normally help distinguish objects from their surroundings are minimal or absent in small objects due to the limited pixels available for analysis. As a result, feature extraction is often restricted to color and context. Multiple small objects with ambiguous features in a large field of view make the object detection task very challenging. Fig. 1 depicts an example of a satellite image that contains small objects of interest (vehicles) highlighted with green boxes. The spatial resolution of this satellite image is 0.3 meters, and each car has about 150 pixels. Although vehicles are recognizable, compared to other ground objects, e.g., buildings, they are much smaller. Portions of large objects may resemble small targets, e.g., objects on roofs, increasing false detections. In addition, small objects contribute little to the overall loss in training, and models tend to favor larger objects.

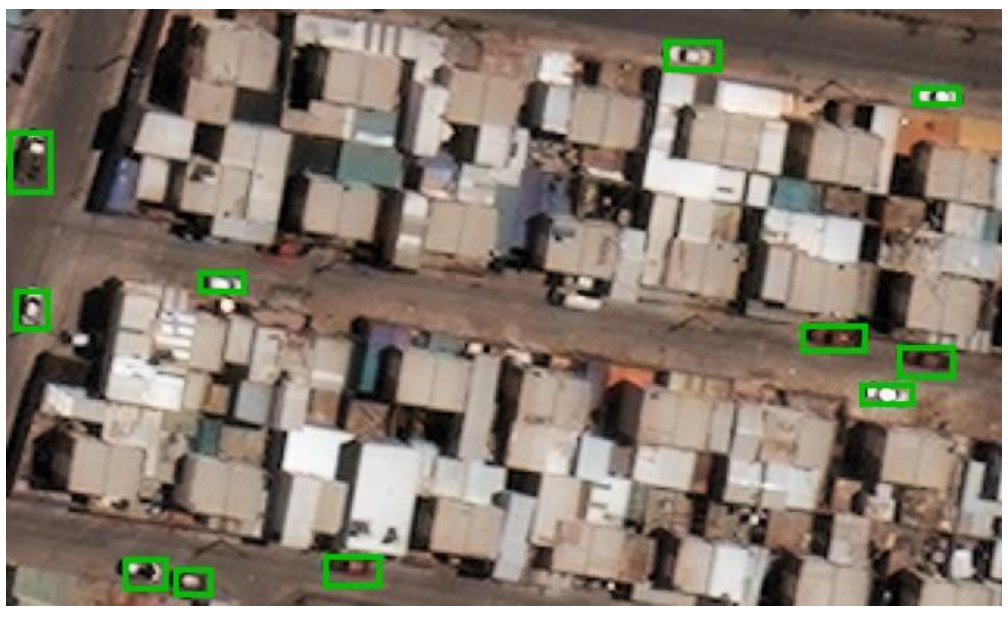

Figure 1: Small objects (in green boxes) in an xView image.

Detecting objects from images and videos is a long-standing computer vision problem, and many methods have been de-veloped in the past decades, from sliding window strategies to deep learning methods. Several literature reviews describe the technical landscape and current challenges [12]. Most recently, Sun et al. [13] reviewed deep learning-based object detection methods, including Convolution Neural Network-based and transformer-based methods. Huang et al. [14] focused on the few-shot and self-supervised learning methods that deal with learning from unlabeled data. Gui et al. [15] reviewed deep learning methods for object detection, emphasizing approaches addressing data and label limitations. Liu et al. [16] surveyed deep learning methods for small object detection from aerial imagery. Recent taxonomic surveys on small object detection with respect to video object tracking [17, 18] highlight their respective challenges, while Rekavandi et al. [11] emphasize task-specific challenges in scenarios of detecting small ships, ranging from dataset acquisition to dynamic reflectance shifts in water. These surveys provide a broad view of the recent technological aspects of object detection methods, as well as their chronological and geographical evolution. Yet, benchmarking of these methods is missing. In particular, evaluating how existing methods perform for the problem of detecting small but critical objects in complex and large views of satellite imagery requires investigation. In addition, small object detection is often complicated by the lack of examples [11].

In this paper, we review object detection methods for finding small objects in remote sensing imagery and provide an empirical evaluation to gain insights into method performance and technical challenges. We use car detection from urban satellite images and honey bee hive detection from satellite images of agricultural lands as application scenarios. Drawing from existing surveys and literature, we identified several top-performing methods for evaluation. Our experiments use two public, high-resolution satellite imagery datasets (xView [19] and SkySat [20] imagery for method evaluation.

The contributions of this work include

- A review of the deep learning methods for the detection of small objects from satellite imagery.
- An in-depth evaluation of the strengths and weaknesses of the state-of-the-art deep learning methods for small object detection.
- A multi-faceted analysis and discussion of the problems and progress, as well as the open challenges.

The remainder of this paper is organized as follows. Section 2 reviews the related work on deep learning methods for object detection with a focus on small object detection. Section 3 presents our empirical study, including evaluation aspects and metrics. Section 4 discusses the results from five major aspects and highlights the advantages and open issues. Section 5 concludes this paper with a summary.

## 2. Related Work

Recent advances in deep networks have produced many object detection methods and corresponding surveys. Liu et al. [16] reviewed small object detection in aerial imagery, noting that shallow networks lack sufficient feature and contextual representation. Kang et al. [21] examined aerial and satellite datasets, emphasizing the impact of spatial resolution on average precision. Li et al. [22] summarized remote sensing detectors and highlighted challenges posed by extreme aspect ratios (e.g., roads, bridges). Amjoud et al. [12] categorized detectors into anchor-based, anchor-free, and transformer-based families, while Li et al. [23] provided a detailed review of transformer detectors, stressing the scalability of Deformable DETR [24]. Zou et al. [25] presented a 20-year historical overview, and Gui et al. [15] underscored the difficulty of building general-purpose models, advocating task-specific designs.

Several surveys also emphasize evaluation. Huang et al. [14] reviewed few-shot and self-supervised detection, observing heavy reliance on pre-trained backbones and benchmarks such as ImageNet and COCO. Cheng et al. [10] benchmarked methods across large-scale datasets, including anchor-free and transformer-based architectures, and noted the difficulty of detecting small, oriented targets in satellite imagery (e.g., DOTA [26]). Sun et al. [13] compared CNN and transformer approaches from region proposal to end-to-end frameworks, but most results are drawn from general-purpose datasets like Pascal VOC, MS COCO, ImageNet, and Cityscapes.

Other studies address persistent small-object challenges. Xie et al. [18] showed that augmentation offers only modest gains due to weak feature representation. Taxonomic surveys on detection and tracking highlight the importance of deep and transfer learning for small objects [17]. Rekavandi et al. [11] benchmarked deep models on objects as small as 16–32 pixels across street-view, MS COCO, and ShipRSImageNet, but their satellite analysis focused mainly on boats, leaving broader land-based evaluations underexplored. These findings reinforce the domain-specific difficulties of satellite imagery.

Table 1: Taxonomy of object detection methods. E2E: End-to-End; Anch: Anchor; RP: Region Proposal; Attn: Attention; MS: multi-scale; Res: Computing resource for training; Spd: Training Speed; Acc: Accuracy; L: Low; M: Medium; H: High.

| Method | Year | Network Structure & Component | | | | | Res | Spd | Acc |
|---|---|---|---|---|---|---|---|---|---|
| | | Stage | Anch. | RP | Attn | MS | | | |
| YOLOv11 [27] | 2024 | One | | | ✓ | ✓ | L | M | H |
| FFCA-YOLO [28] | 2024 | One | | | ✓ | | L | M | H |
| SSD [5] | 2016 | One | ✓ | | | | L | H | L |
| FocusDet [29] | 2024 | One | | | | | M | M | M |
| Faster R-CNN [7] | 2016 | Two | ✓ | ✓ | | ✓ | H | H | H |
| Cascade R-CNN [8] | 2018 | Two | ✓ | ✓ | | ✓ | H | H | H |
| DETR [9] | 2020 | E2E | | | ✓ | ✓ | H | L | M |
| Def. DETR [24] | 2020 | E2E | | | ✓ | ✓ | H | M | H |
| RT-DETR [30] | 2024 | E2E | | | ✓ | ✓ | H | M | H |

Table 1 summarizes the top object detection methods, outlining their architectures, resource requirements, training speed, and accuracy. One-stage detectors such as YOLO and SSD demand fewer computational resources, making them suitable for edge or mobile devices, and many variants specifically target efficient deployment. In contrast, Faster R-CNN and Cascade R-CNN offer strong online processing speed, while YOLO and RT-DETR balance longer training times with competitive inference rates. Given sufficient training data, transformer-based models (e.g., Deformable DETR, RT-DETR) and YOLO methods typically deliver the highest accuracy.

Despite extensive surveys on object detection, few directly address small object detection in remote sensing imagery. Recent attention has shifted to transformers and hybrid CNN-transformer models due to applications such as urban and agricultural analysis. However, satellite imagery introduces sharper performance–accuracy trade-offs [21], exposing the limits of current methods. Moreover, the definition of a "small" object remains inconsistent across studies. This paper fills these gaps by providing a systematic empirical evaluation of state-of-the-art small object detection methods for satellite imagery.

## 3. Small Object Detection and Analysis

### *3.1. Small Object Detection from Remote Sensing Images*

Small objects are often decided by the number of pixels representing the object or by the object size relative to the image size [10, 11]. For example, the MS COCO dataset defines small objects as the ones with less than 32×32 pixels (i.e., about 1,000 pixels) [31], in which the image size is about 640×480. In remote sensing applications, these objects may be represented by one pixel or a small cluster of pixels, depending on the spatial resolution of the satellite imagery. The count of pixels provides a quantifiable means of classifying small objects. Alternatively, a small object can be decided based on the size of the object relative to the image size [32]. A ratio of 0.1 or smaller is often used as the criterion [18]. The rationale for using relative size is rooted in the limited receptive field of deep networks. To meet the network's expectation of the input size, images are downsampled or resized, which reduces the relatively smaller object represented by fewer pixels. Such downsampling also occurs within the network layers via a pooling operation or when the stride of convolutions is greater than one.

Despite numerous recent studies, the definition of small objects remains ambiguous and often subjective. Compared to large objects, the features extracted from small objects are limited, making them prone to misclassification and highly susceptible to noise and distortions. Such problems are particularly evident in remote sensing image analysis, where the vast number and diversity of objects make small targets less significant for detection [11]. In addition, the spatial resolution of remote sensing imagery varies greatly from centimeters to kilometers. Hence, an object may appear large or small in images of different spatial resolutions. Unlike object detection in autonomous driving or action recognition, tiling is used to make images computationally processable. By creating small tiles, we increase the ratio of objects and image sizes, which makes the relative size inappropriate. Therefore, the count of pixels is more suitable for classifying small objects in remote sensing applications.

Fig. 2 illustrates images of xView and SkySat. The spatial resolution of xView and SkySat is 0.3 meters and 0.5 meters, respectively. Both images are rescaled for visualization. The count of pixels of cars is less than 150. There are even smaller objects, less than 10 pixels. However, those objects are hardly recognizable by humans and, hence, are excluded from the detection task. Another type of small object used in this study is honey bee hives, which appear in SkySat images and are represented with bounding boxes consisting of 15 pixels or less.

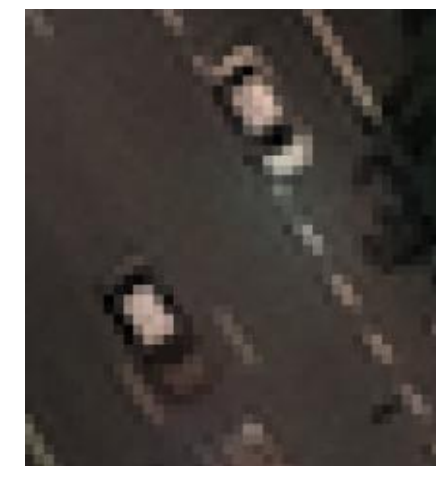

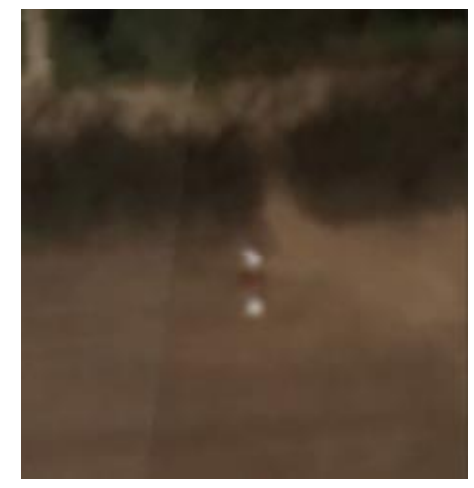

Figure 2: Example of small objects in xView (left) and SkySat (right).

### *3.2. Deep Learning Method Selection*

Based on our review, we identify a group of representative methods for empirical evaluation. Given that most methods were developed for generic object detection tasks instead of targeting small object detection, one criterion of our selection is the key performance factors, including accuracy and robustness. As reported in the surveys, a widely used performance metric is mean average precision (mAP), and the commonly used datasets include VOC (2007 and 2012) and MS COCO.

Another factor we consider in the selection of methods is the network architecture, which could have a significant impact on the detection of small objects in satellite imagery. For instance, the use of high-level features with low spatial resolution often leads to the loss of fine-grained details, and, hence, struggles with small object detection [33]. In general, object detection methods consist of a structured pipeline that transforms an input image into object classifications and localizations. The key components include feature extraction, region proposal, and object classification. In the region proposal, the anchor technique is often used to specify bounding boxes (i.e., anchors), which helps detect objects of varying sizes and aspect ratios by predefining a set of bounding boxes at different locations in an image. Besides region proposal, single-shot methods (a.k.a., one-stage methods) perform object localization and classification in a single pass by predicting bounding boxes and class labels from the input image. However, single-shot methods could result in lower accuracy for detecting small objects.

We identified the following methods that demonstrated superior performance across the aforementioned three datasets [12]: YOLO, SSD, Faster R-CNN, Cascade R-CNN, Real-Time DEtection TRansformer (RT-DETR), deformable DETR, and two methods specifically designed for small object detection: FocusDet [29] and feature enhancement, fusion and context aware YOLO (FFCA-YOLO) [28]. In addition, these methods are representative of feature extraction, region proposal generation, and network structure.

### *3.3. Overview of the Selected Methods*

In R-CNNs, features are extracted from bounding boxes using a CNN, followed by classification to detect objects. Faster R-CNN [34] uses a region proposal network (RPN) that employs a sliding window to generate reference boxes (a.k.a. anchors). It enhances the feature extraction of small, fixed-sized objects. Cascade R-CNN [8] is a multi-stage approach with multiple bounding box regression heads and uses the pooling operations to extract fixed-size feature maps during the refinement process of proposals from regression. The model resamples its proposal distribution for each stage to avoid imbalances between its positive and negative samples at larger threshold values. Cascaded R-CNN balances noise sensitivity and detection degradation through regressors. Each stage consists of a regressor that refines its bounding box on the resampled distribution. The IoU threshold increases per stage on the regressors. This is done to account for alignment and localization degradation observed from iterative regression under higher IoU thresholds. The stage-wise regressor optimization minimizes overfitting and provides more precise bounding boxes.

SSD [5] provides a lightweight backbone-head structure and uses the grid to locate potential objects. Predictions are made at each scale to account for objects of various sizes, where multiscale feature maps are generated using progressively smaller convolutional kernels. Predefined anchor boxes are passed over the image in a sliding window fashion over the grid to establish the position and aspect ratio of the object to the anchor box. This is followed by alignment and localization refinements through center offsets to improve localization. SSD faces challenges in detecting small objects due to the loss of feature information from excessive downscaling of feature maps. Also, having predefined anchor boxes may lead to issues with objects that do not conform to the predefined box.

The YOLO methods frame object detection as a regression problem, using a single convolutional network for object location and prediction. Recent YOLO methods, including YOLOv11 [35], make use of cross-stage attention with center point prediction on objects. These methods optimize cross-stage partial (CSP) connections [36] using convolutions of small kernel size (e.g., 2) for improved gradient flow and reduced computational overhead. Additional integration of features from shallow layers into deeper layers maintains multiscale feature representation during downscaling. This allows for a small receptive field to improve localization in small object detection. Another improvement is the addition of the cross-stage partial with spatial attention (CSPSA) block after the fast spatial pyramid pooling (SPPF) to improve the retention of regions of interest for objects of varying sizes and positions.

RT-DETR [30] uses deformable attention with set hybrid multi-scale interactions to dynamically adjust the receptive field. The intra-scale feature interactions are self-attention-based, while the inter-scale cross interactions are through convolution-based feature fusion. Such inter- and intra-scale interactions benefit in detecting small objects with improved local feature representations and contextual information.

Deformable DETR [24] uses a spatial attention module based on deformable convolutions [37], which dynamically samples each location at different offsets. The decoder consists of multiscale deformable attention and self-attention modules for improved object separation. The multi-scale feature integration through cross-scale interactions aggregates feature maps and enables fine-grained feature extraction.

FocusDet [29] implements bottom focus-PAN that considers the entire feature map for fusion, while the feature enhancement, fusion, and context-aware YOLO (FFCA-YOLO) [28] implements dynamic channel-wise feature re-weighting. Both methods extend YOLOv5.

### 3.4. Empirical Evaluation Method

Our empirical study includes experiments for the use of anchor boxes for region proposal generation, the model architecture, and method-specific techniques. In addition, we evaluate the adaptability of the compared methods when different datasets are used for training and testing the models.

#### 3.4.1. Evaluation Aspects

Anchor boxes, when used in SSD and R-CNNs, involve size and aspect ratio, which decide the range of the detectable object size. A small anchor box allows less distraction from nearby objects, whereas a large anchor box enables the detection of large objects. In our experiments, we evaluate both the size and aspect ratio. We also evaluate the trained models for the detection of medium and large objects. To get the range of the object size, we compute the minimum and maximum sizes of the annotated objects and use them as references for deciding the anchor box size. A factor of two is used to set the range. A similar strategy is used to decide the range. Fig. 3 (a) depicts the histograms of object size (number of pixels) of all classes and of the small objects (cars) in the xView dataset, respectively. The overall distribution of the object size distribution is positively skewed, indicating that most datasets have small objects in bin 6, which represents roughly 403 pixels. The number of object instances that fall under the selection for cars to roughly 5% of the overall size of objects in the dataset. Hence, we focus on keeping the anchor box sizes within 403 pixels. Fig. 3 (b) shows the histogram of the aspect ratio of all objects from the xView dataset. The aspect ratio is computed by dividing the shorter side by the longer side of the anchor box, and its range is (0 1]. Overall, the distribution of the histogram is negatively skewed with a peak at the bin containing aspect ratios in the range of 0.7 to 0.8 and a substantial increase from 0.4. Implying most of the small objects have a slightly rectangular bounding box. We set our anchor boxes to the aspect ratios in the range from 0.7 to 0.8 to better align with the small objects.

Using anchor boxes, region proposals are generated by RPNs in both Faster R-CNN and Cascade R-CNN. The RPNs use thresholds to decide if a proposal is plausible, which influences the quality of the proposals and the number of proposals. A small threshold often yields a large number of proposals, but could produce many weakly supported ones, which contribute to false positive detections. However, a large threshold enforces a strict criterion but will probably lead to missed detections (i.e., false negatives). Our study evaluates the performance of the models using different thresholds. Our region proposal thresholds start at 0.5 with an increment of 0.1 to 0.9.

In transformer-based methods such as YOLOv11, RT-DETR, and deformable DETR, attention is used instead of anchor boxes and region proposals. Without specifying a geometric shape, attention highlights the freeform regions of interest by assigning greater weights to the image pixels of the target objects. Using different parameters (e.g., sampling points and decoder heads), we alter the generation of the attention maps to compare the model performance. In addition, we analyze the behavior of attention through weights and key reference points obtained from the trained models during inference. In particular, we use the number of sampling points as 4, 8, 16, and 32 for deformable DETR and RT-DETR in the model training. A greater number of sampling points could offer an improved detection for small objects. The number of decoder attention heads depends on the complexity of the tasks. In general, a large number of heads offers more aspects and features to capture nuanced relationships. The typical number of heads for transformer-based methods such as deformable DETR and RT-DETR is 8. Increasing the number of heads could lead to an improvement in small object detection because each head captures features from different parts of the image. Hence, we evaluate deformable DETR, RT-DETR, and YOLOv11 using 4, 8, and 16 decoder heads. To isolate the impact of the number of sampling points and decoder heads on the attention, we use 8 sampling points in the evaluation of decoder heads and use 8 decoder heads in the evaluation of the sampling points.

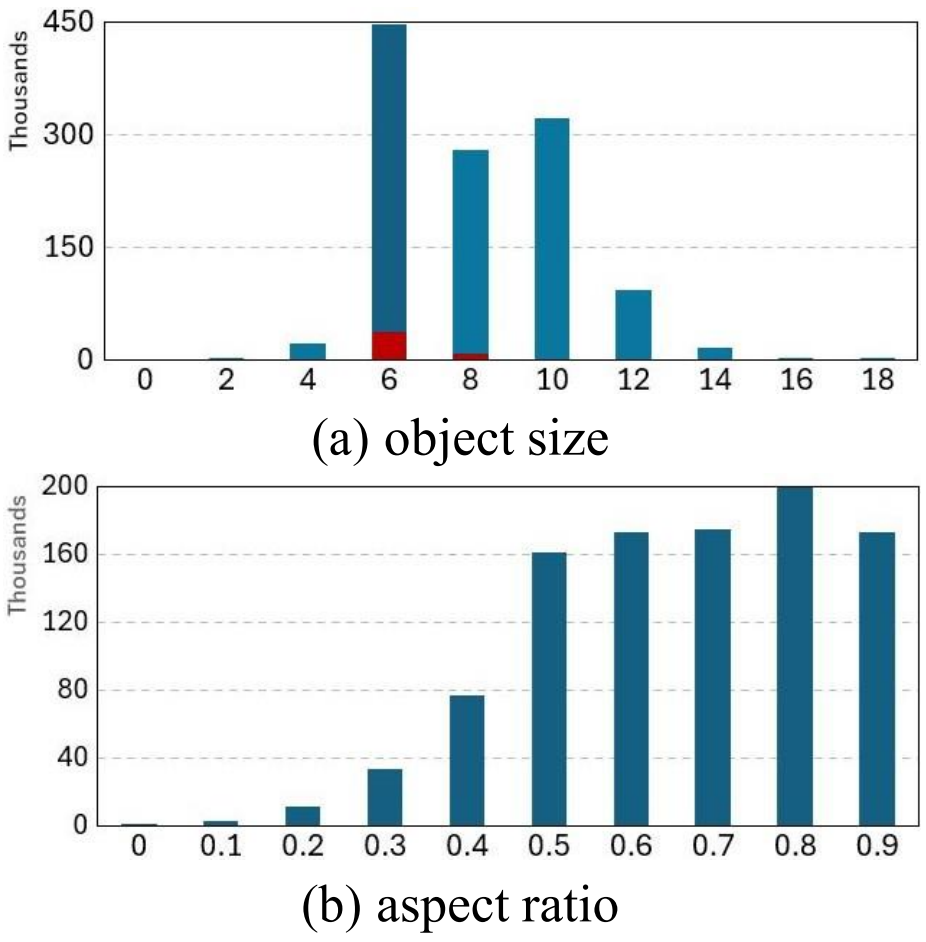

Figure 3: The distribution of object size (number of pixels) and aspect ratio of bounding boxes of the xView dataset. The x-axis in (a) is on a log scale. The blue bars depict the distribution of all objects, and the red bars show the distribution of small objects.

Feature fusion is the process of combining feature representations from different layers to make comprehensive feature maps. An approach used to incorporate feature fusion with multi-scale is the feature pyramid network (FPN) [33] that takes an image at a single scale and generates refined feature maps at different scales. The bottom-up pathway of FPN consists of forward-passing convolutional layers that downscale feature maps to obtain high-level semantic information. The top-down pathway upsamples the higher-level feature maps in the pyramid to the same scales as the bottom-up pathway with the nearest neighbor algorithm. The lateral connections fuse the feature maps from the bottom-up and top-down pathways at respective scales on the same level. This is achieved using a $1\times1$ convolution for channel reduction and element-wise addition, which helps the lower-level maps compensate for the lack of semantic information from the higher levels obtained from the top-down pathway. We evaluate multi-scale and feature fusion for Faster R-CNN, Cascade R-CNN, YOLOv11, deformable DETR, and RT-DETR by training the methods with and without the FPN module in the backbone and using single feature maps at size $512\times512$ pixels. Removing the FPN module in the backbone of the method architectures removes multi-scale feature representations during the feature extraction process into a single feature map from the convolutional networks. Faster R-CNN and Cascade R-CNN are trained for 5,000 iterations. YOLOv11, deformable DETR, and RT-DETR are trained for 28,320, 56,580, and 75,450 iterations, respectively.

Generalization of the methods for small objects is evaluated by fine-tuning the trained models through transfer learning. The source dataset is the xView trained on 75% of the data. The target dataset is the SkySat, with 80% of the images used for fine-tuning the models and the remaining for validation. The dataset is evaluated with 6-fold cross-validation and setting hyperparameters from Table 5. We change the number of iterations of YOLOv11, SSD, RT-DETR, Faster R-CNN, Cascade R-CNN, and deformable DETR to 1477, 10,000, 9,500, 5,000, 5,000, and 6875, respectively. Due to the limited number of training samples of SkySat, we freeze the backbone layers. We evaluate the generalization ability by comparing the number of detected objects against the ground truth and the overlap of the bounding boxes with the targets. The SkySat images usually have a lower resolution and fewer bee hives in rural settings. However, there are cases, such as bee farms and temporary holding fields, that have large clusters of hives. These objects have varying shapes and intensities, making it challenging to derive distinct patterns that distinguish them from the surroundings.

### 3.4.2. Datasets

In this study, we conduct evaluations using three datasets: satellite imagery datasets: xView [19], SkySat [20], and Dataset of Object deTection in Aerial images (DOTA) [38]. Both are high-resolution satellite imagery. Table 2 summarizes the key properties of the two datasets.

Table 2: Data properties. Channels are CB: Coastal Blue; R: Red; G: Green; B: Blue; Y: Yellow; RE: Red Edge; NIR: Near Infrared; Pan: Panchromatic.

| Dataset | Spatial Resolution (m) | Channels | Area coverage (km) |
|---|---|---|---|
| xView | 0.3 | CB, B, G, Y, R, RE, NIR1, NIR2 | 1,400 |
| SkySat | 0.5 | B, G, R, NIR, Pan | 66 |
| DOTA | 0.01-3.2 | B, G, R | varied |

xView provides a publicly accessible satellite imagery dataset capturing diverse scenes with a spatial resolution of 0.3 meters and 8-channel multispectral satellite images for object detection. The images include objects of 60 classes in different sizes, from which we select the ones that fit our definition of small objects based on the number of pixels of the largest objects in that class. The number of pixels is limited to fewer than 1,000. An example is shown in the left panel of Fig. 2. SkySat offers 0.5-meter spatial resolution, 4-channel satellite images with high temporal coverage. Besides urban coverage, it provides a wide coverage of rural lands. One of the unique objects in rural lands is honey bee hives, which are typically small, with a size of approximately 15 pixels in the SkySat images. DOTA consists of a variety of aerial images having different spatial resolutions for their satellite imagery. Small vehicles are the smallest and most prevalent class within the dataset, with 242,276 instances having object sizes within 1,000 pixels. Table 3 reports the size and distribution of the four classes of the xView dataset, which consists of objects with a size less than 1,000 pixels. These classes include Small Car, Passenger Vehicle, Pickup Truck, and Utility Truck. The Small Car class is a generic class that includes specific vehicle types but is not clearly differentiable. The number of images is the count of images that contain at least one object of the class. In many cases, an image contains small objects of more than one class. Hence, the total number of images of all classes is greater than the number of xView images listed in Table 2. The number of objects is the count of the small objects of each class in all images. Object size lists the minimum, maximum, and median size (number of pixels) of the objects in each class. It shows the Small Car class has the largest dynamic range in size.

Table 3: Size and distribution of classes in the xView dataset that meet the criterion of small objects.

| Classes of small objects | Num. of Image | Num. of Objects | Object size (pixels) | | |
|---|---|---|---|---|---|
| | | | min | median | max |
| Small Car | 691 | 423,092 | 20 | 130 | 957 |
| Passenger Vehicle | 203 | 5,900 | 20 | 130 | 735 |
| Pickup Truck | 207 | 2,008 | 102 | 180 | 700 |
| Utility Truck | 354 | 7,264 | 14 | 198 | 840 |

### 3.4.3. Evaluation Metrics

Determining the right metrics for evaluating these methods requires consideration of how well the bounding boxes localize on the object and provide the correct classification. Average Precision (AP) captures the precision of object detection across a range of confidence thresholds. As a model outputs predicted bounding boxes with class labels and confidence scores, AP measures how well these predictions match the ground-truth.

Fig. 4 illustrates how the IoU threshold at 50% considers evaluating detected bounding boxes. Consider an image consisting of five small vehicles, annotated with bounding boxes in green. The object detection method predicts multiple bounding boxes, along with their respective confidence values, which range from 0 to 1. Fig. 4 (a) depicts 13 bounding boxes sorted in descending order of confidence along with their respective IoU scores. The IoU threshold is used to decide how well a prediction aligns with the ground truth. A prediction is considered a true positive (TP) if it has the correct class label and overlaps with the ground-truth (measured with IoU) above a threshold represented in red. If multiple bounding boxes meet the criteria for a single ground truth, then the highest confidence score detection is marked as TP. This enforces penalization for duplicate bounding boxes per ground truth. Predictions that fail to match any ground truth or are duplicate detections of the same object are counted as false positives (FP), shown in blue. False negatives (FN) are obtained from misses in detection, where there is no overlap between the predictions and ground truths. A precision-recall (PR) curve is plotted through different confidence values. An AP score across different thresholds is weighted with recall to account for the number of detections with precision. The AP score at an IoU threshold $t$, denoted with $AP_t$, is calculated based on the area under the curve through Riemann summation:

$$AP_t = \sum_{i=1}^{M} \Delta R_t^i \max_{j \geq i} P_t^j, \tag{1}$$

where $P_t = \frac{TP_t}{TP_t + FP_t}$, and $R_t = \frac{TP_t}{TP_t + FN_t}$, are precision and recall at a threshold $t$, respectively. $\Delta R_t^i = R_t^i - R_t^{i-1}$, is the change.

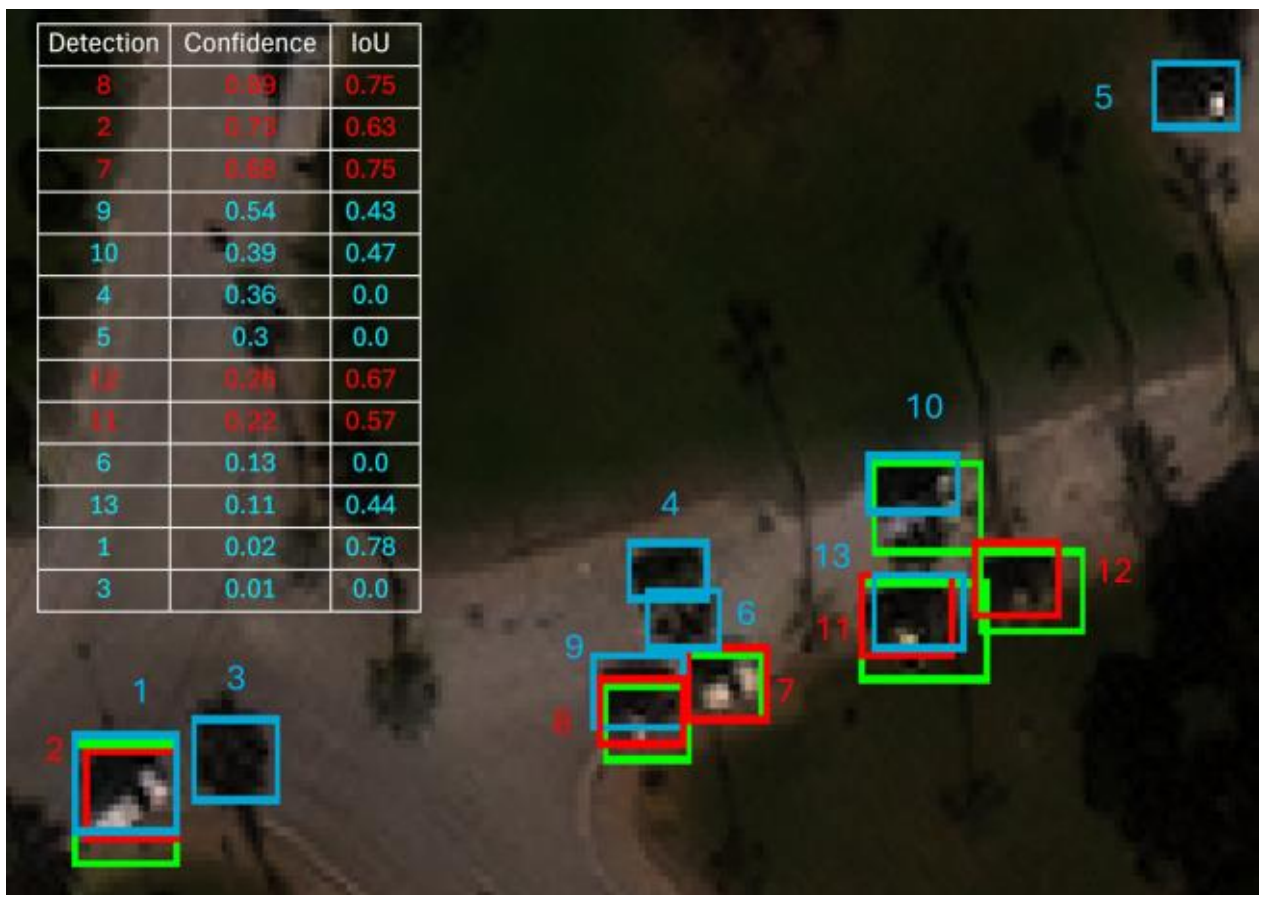

(a) True positive detections

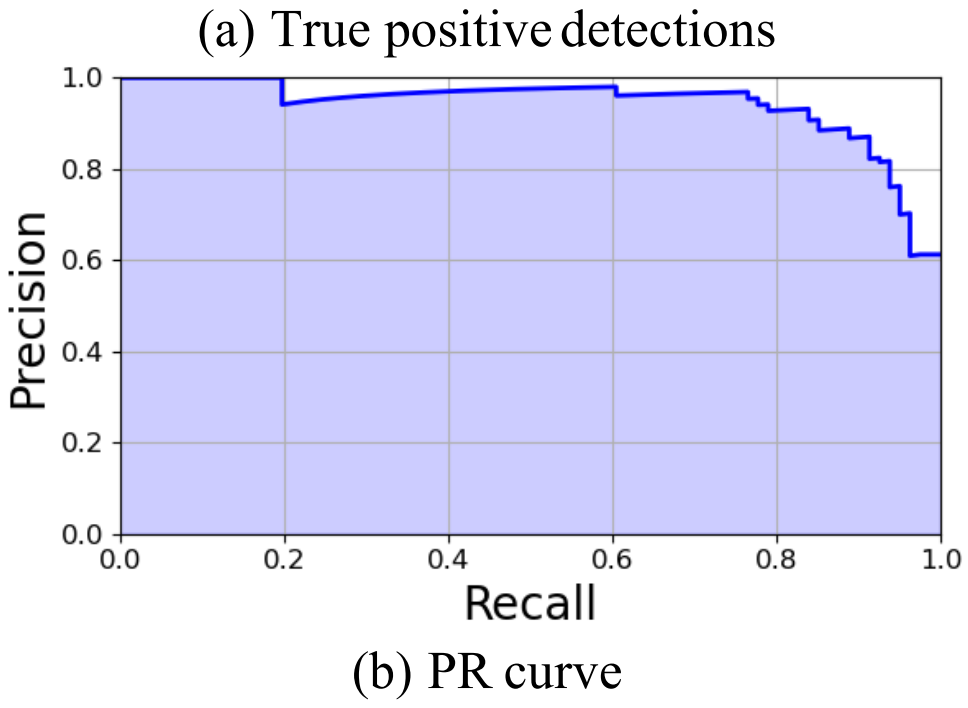

(b) PR curve

Figure 4: AP/AR calculation at threshold 50. (a) Detected small objects, where GT: green, TP: red, FP: cerulean. The confidence is shown on the left. (b) PR curve in blue with filled area underneath.

The area under the PR curve, such as in Fig. 4(b), represents the trade-off between precision and recall across all confidence thresholds. A large AP indicates that the model accurately and consistently detects objects across various sizes and positions, with fewer FPs and missed detections. When multiple IoU thresholds are used, e.g., from 0.50 to 0.95 in an increment of

0.05, AP is computed as follows:

$$AP_{\tilde{t}} = \frac{1}{10}\sum_{t=0.5}^{0.95} AP_t, \tag{2}$$

where $\tilde{t}$ indicates a threshold range, which varies from 0.50 to 0.95 with a step size of 0.05 across 10 thresholds. Fig. 5 illustrates the categorization of TPs and FPs from the detected bounding boxes across different IoU thresholds. As the IoU threshold increases, the condition for TPs becomes stricter, resulting in fewer TPs and an increase in FP. At the threshold of 75, only two bounding boxes meet the requirement (displayed in red), while at the threshold of 95, all boxes fail to meet the criterion. $AP_{50:95}$ emphasizes balance and consistency.

Fig. 5 presents the precision-recall curves for the example image across different IoU thresholds ranging from 0.50 to 0.95 in increments of 0.05. Solid lines represent the scores of precision and recall, ranging from 0 to 1.0. As the IoU threshold becomes stricter, the number of TP reduces, leading to drops in both precision and recall limits. The decrease in precision and recall scores is also visible with the shrinking of the PR curve at the higher thresholds. In this scenario, any threshold higher than 0.85 has no bounding boxes satisfying the criteria to be a TP and labeled as FPs. Additionally, the lack of TP around ground truths increases FNs, thereby reducing recall scores. While the lower threshold values yield higher AP scores due to more lenient matching, calculating the average across all thresholds provides a more comprehensive score that accounts for imbalances in the overall distribution of scores.

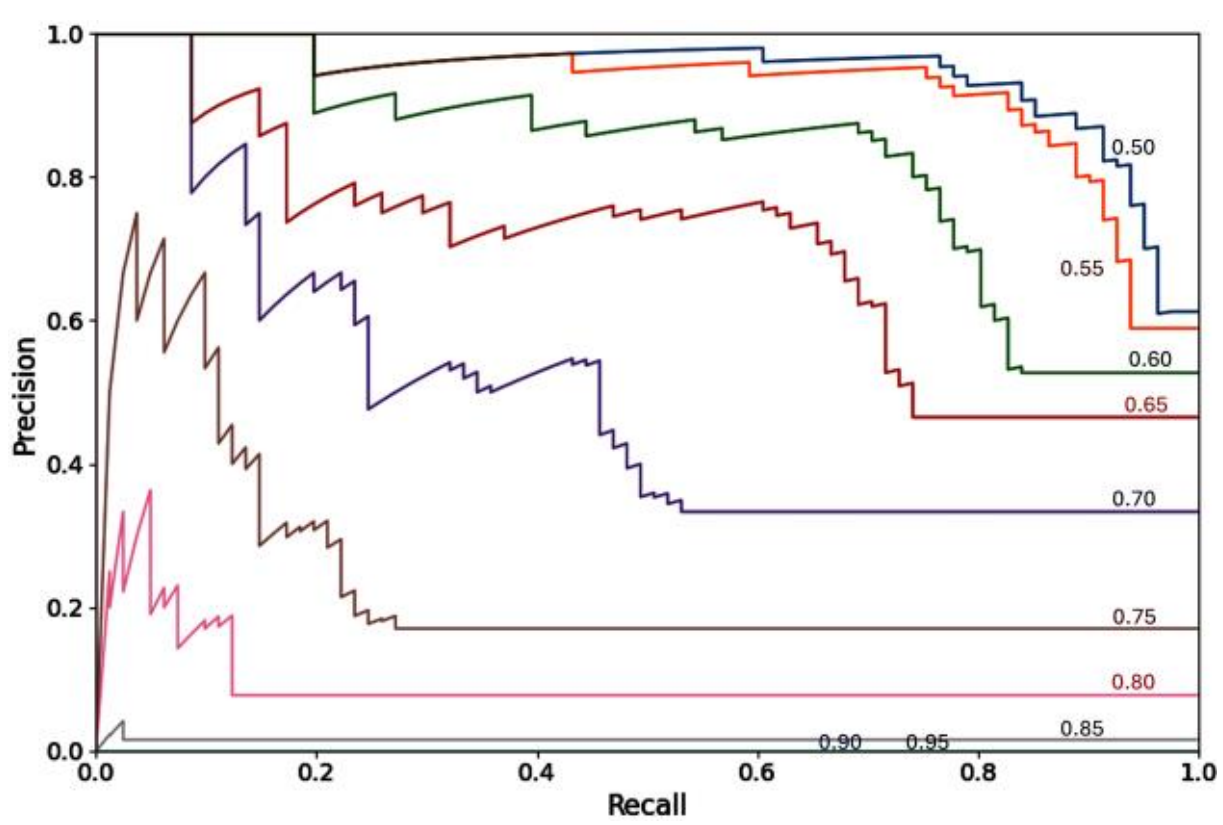

Figure 5: PR Curve at IoU thresholds from 0.50 to 0.95 in steps of 0.05.

$AP_{50}$ signifies at least 50% overlap with the ground truth. While $AP_{50:95}$ provides an integrative view of detection across a range of IoU thresholds, $AP_{50}$ highlights the trade-off between sensitivity and localization. For example, a high $AP_{50}$ with a low $AP_{50:95}$ indicates that the method struggles to localize its detections accurately. Conversely, a low $AP_{50}$ with a high $AP_{50:95}$ indicates fewer detections with more precise bounding boxes. Similarly, we report the average recall (AR) as follows:

$$AR_t = \frac{1}{N}\sum_{c=1}^{N} R_{t,c}^{k}, \quad \text{and} \quad AR_{\tilde{t}} = \frac{1}{10}\sum_{t=0.5}^{0.95} AR_t, \tag{3}$$

where $N$ represents the number of classes at threshold t and top k (k=100) proposals.

Besides AP and AR metrics, we also report the $F1$ score that computes the harmonic mean of precision and recall over a range of IoU thresholds as follows:

$$F1_{\tilde{t}} = \frac{1}{10}\sum_{t=0.5}^{0.95} \frac{2P_t R_t}{P_t + R_t}. \tag{4}$$

## 4. Results and Discussion

### 4.1. Experimental Settings

The raw images from the xView and SkySat are cropped into tiles (images) of size 512×512. If the image size is a multiple of 512 in both height and width, the tiles are cropped without overlaps. Otherwise, we compute the number of tiles by rounding the division of the height (or width) by 512 to the larger integer and perform cropping accordingly. Hence, overlaps between tiles are allowed. This ensures the complete coverage of the image by the cropped tiles with a relatively small overlap. For each dataset, the processed images are split into two parts, with 70% for training and the rest for validation. Table 4 lists the volume of cropped images. We conducted 6-fold cross-validation and report the average performance in our experiments. The number of training images and validation images is per fold in cross-validation. Image augmentation is applied through random flipping with a probability of 50%.

Table 4: Experimental data for training and validation.

| Dataset | Number of Images | | Number of Images | |
|---|---|---|---|---|
| | Raw | Processed | Training | Validation |
| xView | 1,413 | 9,054 | 7,545 | 1,509 |
| SkySat | 34 | 66 | 55 | 11 |
| DOTA | 1,869 | 7,148 | 4,765 | 2,383 |

SSD is trained for 20,000 iterations with learning steps at 6,000 and 8,000. It has seven layers, with a spatial resolution of the feature maps decreasing by a factor of 2 from $2^6$. The stride per layer is calculated by dividing the image size by the size of the feature map at each layer. Faster R-CNN and Cascade R-CNN have feature pyramid network modules with multi-scale incorporated to enhance the feature maps extracted by the backbone. The region proposal threshold is 0.5, and anchors equal to or greater than the threshold are considered to contain valid objects. Faster R-CNN is trained for 10,000 iterations, and Cascade R-CNN is trained for 20,000 iterations. FFCA-YOLO was trained for 100 epochs, and FocusDet was trained for 200 epochs. Both methods were trained on the xView dataset with a batch size of 16.

Optuna [39] is used to fine-tune the models for small object detection through Bayesian optimization. Table 5 summarizes the hyperparameters of the compared methods after fine-tuning. The first two rows are the lower and upper limits of the search space in the tuning process. The dashes indicate non-existent

$c$=1 $t$=0.5

Table 5: Network hyperparameters after fine-tuning using Optuna. Learning Rate (LR): $10^{-4}$; Batch Size (BS); Range of Anchor Box Size (RABS) starting from 10; number of queries (NQ); Decay: $10^{-4}$; Iteration (Itr.): $10^3$.

| Method | LR | BS | RABS | NQ | Decay | Mom. | Itr. |
|---|---|---|---|---|---|---|---|
| Range (min) | 0.1 | 1 | 10 | 50 | 0.001 | 0.7 | - |
| Range (max) | 100 | 32 | 400 | 400 | 100 | 0.99 | - |
| SSD | 7.00 | 32 | 300 | - | 0.139 | 0.8 | 20 |
| Fas. R-CNN | 38.10 | 16 | 90 | - | 0.0285 | 0.90 | 270 |
| Cas. R-CNN | 91.00 | 16 | 90 | - | 0.0279 | 0.90 | 270 |
| YOLOv11 | 1.01 | 8 | - | - | 5 | 0.89 | 94 |
| RT-DETR | 7.68 | 3 | - | 300 | 9.88 | 0.92 | 252 |
| Def. DETR | 1.19 | 4 | - | 228 | 55.70 | 0.90 | 252 |
| FFCA-YOLO | 100 | 16 | - | - | 50 | 0.97 | 29 |
| FocusDet | 100 | 16 | - | - | 50 | 0.97 | 59 |

parameters for the methods. For example, YOLOv11 determines the number of queries based on the feature maps; hence, the number of queries is omitted. Labelstudio [40] is used for annotating the SkySat dataset. The SSD and YOLO methods use VGG-16 and C3k2 as backbones, respectively, while the rest use ResNet-50 as the backbone.

### 4.2. *Anchor Box Size*

Among the anchor-based methods, Faster R-CNN and Cascade R-CNN utilize anchor boxes to generate region proposals through RPN modules for object detection. The RPN creates an anchor box for each aspect ratio at each location and scale of the feature maps. Similarly, SSD generates an anchor box of an aspect ratio across each feature map.

Table 6 reports the average AP, AR, and F1 of various aspect ratios at an anchor box size of 90, together with the standard deviation in parentheses. Following the distribution of aspect ratio shown in Fig. 3(b), we evaluate the methods using aspect ratios from 0.55 to 0.95. We use a set of three aspect ratios to create anchor boxes that balance between coverage in capturing diverse shapes and additional computational overhead. For example, in the case of 0.65±0.1, three aspect ratios are used: 0.55, 0.65, and 0.75. We calculate the AP, AR, and F1 using the final bounding boxes obtained from the anchor-based methods. The best and second-best scores are highlighted in boldface and underline, respectively. In addition, the best scores from each method in italics provide further detail.

For each method, aspect ratios in the range of 0.75 to 0.95, i.e., 0.85±0.1, yield the best performance. This is quite consistent for all three methods. However, the performance difference is small, especially for Faster R-CNN and Cascade R-CNN. This could be attributed to the regression head used in both methods. SSD, on the other hand, is more influenced by different aspect ratios. While 0.75±0.1 yields higher AR scores, the slightly lower AP score indicates a greater number of false positives from the background classes. Additionally, the F1 score disparity in SSD is more pronounced with greater sensitivity, portraying a lower balance of precision and recall at lower aspect ratios. In the rest of our experiments, aspect ratios of 0.75, 0.85, and 0.95 are used to maintain consistent performance across the anchor-based methods.

Table 6: AP, AR, and F1 using different aspect ratios.

| Method | Aspect ratio | 50 | | | 50:95 | | |
|---|---|---|---|---|---|---|---|
| | | AP | AR | F1 | AP | AR | F1 |
| Faster R-CNN | 0.65 | 57.40 | 70.74 | ***66.42*** | 21.27 | 32.51 | ***31.13*** |
| | ±0.1 | (2.57) | (3.95) | (1.42) | (0.80) | (1.54) | (0.50) |
| | 0.75 | 56.57 | 71.38 | 65.77 | 20.73 | 32.52 | 30.58 |
| | ±0.1 | (1.87) | (2.61) | (0.66) | (0.42) | (0.73) | (0.37) |
| | 0.85 | ***57.57*** | ***72.27*** | 65.88 | ***21.97*** | ***33.16*** | 30.84 |
| | ±0.1 | (1.51) | (2.70) | (0.53) | (0.12) | (0.61) | (0.41) |
| Cascade R-CNN | 0.65 | 51.90 | 67.94 | 59.83 | *19.90* | 31.58 | *28.89* |
| | ±0.1 | (1.77) | (2.70) | (1.13) | (0.67) | (1.13) | (0.44) |
| | 0.75 | 51.83 | 67.87 | *59.94* | 19.33 | *32.28* | 28.88 |
| | ±0.1 | (1.96) | (2.40) | (1.18) | (0.74) | (0.28) | (0.38) |
| | 0.85 | *52.80* | *68.02* | 59.65 | 19.90 | 31.59 | 28.74 |
| | ±0.1 | (1.95) | (2.60) | (1.28) | (0.78) | (1.09) | (0.51) |
| SSD | 0.65 | 7.67 | 15.97 | 16.00 | 1.93 | 4.03 | 5.67 |
| | ±0.1 | (5.42) | (11.20) | (11.30) | (1.37) | (3.43) | (4.01) |
| | 0.75 | 9.27 | *24.09* | 21.10 | 2.23 | *7.93* | 7.07 |
| | ±0.1 | (2.23) | (3.86) | (3.33) | (0.61) | (1.48) | (1.29) |
| | 0.85 | *9.63* | 21.22 | *22.49* | *2.43* | 7.13 | *7.87* |
| | ±0.1 | (3.08) | (6.56) | (0.14) | (0.84) | (2.50) | (0.12) |

Table 7 reports the AP, AR, and F1 of the three methods using different anchor box sizes. The max size represents the upper limit of the anchor box size. In our experiments, the anchor box size varies from 10 to the upper limit. We conduct 6-fold cross-validation and report the average and standard deviation. As shown in the table, Faster R-CNN and Cascade R-CNN yield better results with a smaller anchor box size at 90, whereas SSD achieves better performance when using a larger anchor box size (300 or 400). However, the performance of SSD is significantly worse than the two R-CNN methods, and the average AP, AR, and F1 are less than one-fifth of those of the R-CNNs. Despite smaller anchor boxes often achieving a better performance, the differences are small.

The significant disparity of $AP_{50}$ and $AP_{50:95}$ indicates that all three methods face localization and misalignment issues. While the R-CNN methods are more consistent with localization, there is a drop in AP from 300 to 400 due to the number of objects with sizes greater than 300 being in the minority, as shown in Fig. 3(a). Also, large anchor boxes often lead to a loose fit to the objects, resulting in decreased performance. Their F1 scores are less than ARs, indicating a higher rate of false positives, which in turn causes lower precision. While SSD has consistent detections at a lower anchor box size of 90, the rate of increase of AR is greater than F1, indicating a more significant impact from misalignment and false positives.

### 4.3. *Region Proposal*

Table 8 reports the average AP, AR, and F1 of Faster R-CNN and Cascade R-CNN by varying the threshold for the region proposal network from 0.5 to 0.9. Both Faster R-CNN and Cascade R-CNN achieve their best performance at a threshold of 0.5, which generates more positive proposals. Compared to Cascade R-CNN, Faster R-CNN exhibits better performance with an $AP_{50}$ of 61.65 and $AR_{50}$ of 73.99, which is about 10% and 5% improvement, respectively.

Table 7: AP, AR, and F1 using different anchor box sizes.

| Method | Max size | 50 AP | 50 AR | 50 F1 | 50:95 AP | 50:95 AR | 50:95 F1 |
|---|---|---|---|---|---|---|---|
| Faster R-CNN | 90 | ***61.65*** | 73.99 | ***69.44*** | ***23.32*** | ***35.78*** | ***32.97*** |
| | | (3.22) | (3.09) | (2.07) | (1.30) | (1.77) | (1.09) |
| | 200 | 60.48 | ***74.33*** | 68.26 | 22.05 | 34.80 | 31.51 |
| | | (2.75) | (3.45) | (1.69) | (1.08) | (1.59) | (0.88) |
| | 300 | 60.75 | 74.06 | 68.55 | 22.25 | 34.95 | 31.86 |
| | | (3.02) | (3.00) | (1.81) | (1.28) | (1.51) | (1.05) |
| | 400 | 59.83 | 74.11 | 68.08 | 21.80 | 34.90 | 31.56 |
| | | (2.94) | (2.97) | (1.97) | (1.04) | (1.49) | (0.93) |
| Cascade R-CNN | 90 | *55.60* | *70.31* | *63.22* | *21.53* | *33.93* | *30.71* |
| | | (2.12) | (3.09) | (1.29) | (0.90) | (1.55) | (0.68) |
| | 200 | 53.93 | 69.63 | 62.12 | 20.07 | 32.33 | 29.24 |
| | | (2.29) | (3.15) | (1.21) | (0.90) | (1.56) | (0.59) |
| | 300 | 53.80 | 69.81 | 62.05 | 20.10 | 32.33 | 29.19 |
| | | (2.41) | (3.14) | (1.43) | (1.11) | (1.56) | (0.70) |
| | 400 | 53.53 | 69.45 | 61.56 | 19.87 | 32.03 | 28.89 |
| | | (2.46) | (3.25) | (1.59) | (1.02) | (1.71) | (0.83) |
| SSD | 90 | 10.43 | 22.31 | 22.45 | 2.58 | 7.52 | 7.69 |
| | | (1.95) | (3.72) | (4.95) | (0.45) | (1.11) | (1.49) |
| | 200 | 10.00 | 23.46 | 21.67 | 2.43 | 7.95 | 7.35 |
| | | (2.26) | (4.44) | (3.76) | (0.51) | (1.43) | (1.17) |
| | 300 | 13.15 | 27.61 | 25.89 | *3.35* | 9.07 | 9.16 |
| | | (1.37) | (2.66) | (2.62) | (0.30) | (0.99) | (0.94) |
| | 400 | *13.35* | *29.09* | *26.85* | 3.28 | *10.17* | *9.41* |
| | | (2.75) | (4.49) | (3.42) | (0.71) | (1.63) | (1.27) |

Table 8: AP, AR, and F1 using different thresholds for region proposal.

| Method | Thres. | 50 AP | 50 AR | 50 F1 | 50:95 AP | 50:95 AR | 50:95 F1 |
|---|---|---|---|---|---|---|---|
| Faster R-CNN | 0.5 | ***61.65*** | ***73.99*** | ***69.44*** | ***23.32*** | ***35.78*** | ***32.97*** |
| | | (3.22) | (3.09) | (2.07) | (1.30) | (1.77) | (1.09) |
| | 0.6 | 57.87 | 72.17 | 68.01 | 21.73 | 33.67 | 32.17 |
| | | (3.16) | (2.69) | (1.72) | (1.10) | (1.18) | (0.75) |
| | 0.7 | 58.10 | 72.18 | 68.10 | 21.97 | 33.67 | 32.24 |
| | | (2.31) | (2.84) | (1.19) | (0.78) | (1.13) | (0.39) |
| | 0.8 | 57.10 | 71.48 | 67.93 | 20.67 | 32.23 | 32.17 |
| | | (2.74) | (2.37) | (1.40) | (0.78) | (0.78) | (0.51) |
| | 0.9 | 56.10 | 71.56 | 68.18 | 19.97 | 32.07 | 32.37 |
| | | (1.91) | (2.55) | (1.38) | (0.85) | (1.18) | (0.70) |
| Cascade R-CNN | 0.5 | *55.60* | *70.31* | *63.22* | *21.53* | 33.93 | *30.71* |
| | | (2.12) | (3.09) | (2.07) | (0.90) | (1.55) | (0.68) |
| | 0.6 | 54.90 | 70.33 | 56.84 | 21.47 | 33.27 | 25.16 |
| | | (1.96) | (3.04) | (1.33) | (0.86) | (1.39) | (0.81) |
| | 0.7 | 54.83 | 70.11 | 56.14 | 21.33 | 33.03 | 24.72 |
| | | (2.08) | (2.92) | (1.54) | (0.90) | (1.35) | (1.07) |
| | 0.8 | 53.63 | 69.76 | 61.59 | 20.70 | 32.57 | 29.85 |
| | | (3.32) | (3.87) | (2.73) | (1.42) | (1.95) | (1.40) |
| | 0.9 | 52.90 | 68.91 | 60.64 | 20.40 | 32.20 | 29.33 |
| | | (3.34) | (3.76) | (2.78) | (1.61) | (1.98) | (1.51) |

A large RPN threshold often results in fewer proposed regions that satisfy the criterion, thereby reducing the number of positives and lowering classification accuracy. This is exhibited by the discernible decrease in $AP_{50}$ and $AR_{50}$ from Faster R-CNN as the threshold increases from 0.5 to 0.9. However, the drop in performance is less significant in $AP_{50:95}$ and $AR_{50:95}$ compared to $AP_{50}$ and $AR_{50}$. While $AP_{50}$ and $AR_{50}$ decrease when larger thresholds are used, $AP_{50:95}$ and $AR_{50:95}$ remain consistent, indicating a shift towards the more localized predictions with large thresholds. Faster R-CNN achieves a better F1 score compared to Cascade R-CNN. However, at a stricter RPN threshold of 0.9, AR50:95 drops below $F1_{50:95}$, indicating weaker localization compared to lower thresholds. Faster R-CNN has more consistent F1 compared to Cascade R-CNN due to proportional linear shifts in precision and recall.

Cascade R-CNN is more consistent than Faster R-CNN, as its performance drop at larger thresholds is negligible, and it achieves the second-best $AR_{50:95}$ at a threshold of 0.5. This can be attributed to the cascaded modules that refine across multiple thresholds. The cascaded decoders used for refinement also inject more hypotheses through resampling at each head based on the refined proposals. Such refinement also leads to uneven PR curves and inconsistencies in the F1 score. Overall, restricting the region proposal network to more localized proposals hinders the learning process by ignoring partial proposals.

### 4.4. Attention Mechanism

Deformable DETR, RT-DETR, and YOLOv11 leverage attention with variable and fixed numbers of sampling points. Table 9 reports the average AP and AR of deformable DETR and RT-DETR at different sampling points. Both methods use eight decoder attention heads. The influence of the number of sampling points is inconsistent between the two methods. RT-DETR is more sensitive to the number of sampling points due to its receptive field coverage across multiple decoder layers. The wider receptive field allows the method to detect more objects, resulting in better AR compared to deformable DETR. When the number of sampling points is 8, both methods produce better results across AP, AR, and F1. RT-DETR achieves the best and second-best results at 8 and 4, respectively. Yet, increasing the number of sampling points incurs more computational overhead without additional benefits and a drop in F1 scores.

Table 9: AP, AR, and F1 using different numbers of sampling points.

| Method | Samp. pts | 50 AP | 50 AR | 50 F1 | 50:95 AP | 50:95 AR | 50:95 F1 |
|---|---|---|---|---|---|---|---|
| RT-DETR | 4 | 67.47 | 77.90 | *77.22* | 27.65 | 38.80 | 36.07 |
| | | (2.91) | (3.40) | (1.74) | (1.26) | (1.72) | (0.97) |
| | 8 | ***68.20*** | ***78.31*** | 72.97 | ***28.15*** | ***39.22*** | *36.48* |
| | | (3.06) | (3.44) | (1.85) | (1.14) | (1.77) | (1.04) |
| | 16 | 61.67 | 73.07 | 67.84 | 24.24 | 34.97 | 32.94 |
| | | (2.13) | (2.96) | (0.93) | (0.88) | (1.11) | (0.35) |
| | 32 | 64.90 | 75.58 | 70.46 | 26.57 | 37.57 | 34.93 |
| | | (3.51) | (3.68) | (2.27) | (1.61) | (1.89) | (1.24) |
| Def. DETR | 4 | 50.70 | *75.48* | ***86.39*** | 18.90 | *36.52* | ***46.71*** |
| | | (2.55) | (3.48) | (0.66) | (1.13) | (1.63) | (0.66) |
| | 8 | 62.20 | 74.35 | 85.79 | 23.43 | 36.45 | 46.09 |
| | | (2.72) | (2.93) | (0.20) | (1.05) | (1.55) | (0.27) |
| | 16 | 60.13 | 75.28 | 86.14 | 22.60 | 35.27 | 46.33 |
| | | (2.55) | (1.57) | (0.52) | (1.12) | (1.60) | (0.36) |
| | 32 | *62.32* | 75.28 | 86.36 | *23.52* | 36.42 | 46.54 |
| | | (2.87) | (3.48) | (0.57) | (1.08) | (1.64) | (0.59) |

The number of decoder heads is another factor of attention. Table 10 reports the average AP, AR, and F1 by changing the number of heads while keeping the number of sampling points at 8. Unlike DETR methods, YOLOv11 uses fixed grid-based sampling from each attention head to process queries. Hence, YOLOv11 is not included in Table 9.

Table 10: AP, AR, and F1 using different numbers of decoder attention heads.

| Method | # of Heads | 50 | | | 50:95 | | |
|---|---|---|---|---|---|---|---|
| | | AP | AR | F1 | AP | AR | F1 |
| RT-DETR | 4 | 63.67 | 74.79 | 69.42 | 25.63 | 36.73 | 34.13 |
| | | (3.77) | (4.15) | (2.68) | (2.08) | (2.39) | (1.72) |
| | 8 | ***68.20*** | ***78.31*** | *72.97* | *28.15* | **39.22** | *36.48* |
| | | (3.06) | (3.44) | (1.85) | (1.14) | (1.77) | (1.04) |
| | 16 | 58.03 | 72.04 | 65.44 | 22.53 | 34.80 | 31.44 |
| | | (6.19) | (4.98) | (3.97) | (3.10) | (2.82) | (2.46) |
| Def. DETR | 4 | 60.33 | 73.12 | 85.89 | 22.70 | 35.13 | ***46.16*** |
| | | (2.64) | (3.31) | (0.55) | (1.07) | (1.45) | (0.40) |
| | 8 | *62.20* | *74.35* | 85.79 | *23.43* | *36.45* | 46.09 |
| | | (2.72) | (2.93) | (0.20) | (1.05) | (1.55) | (0.27) |
| | 16 | 60.53 | 73.42 | ***86.03*** | 22.77 | 35.30 | 46.14 |
| | | (2.68) | (3.28) | (0.50) | (1.22) | (1.50) | (0.36) |
| YOLO v11 | 4 | *67.33* | *76.54* | 72.56 | 28.53 | 39.07 | 36.73 |
| | | (2.65) | (3.39) | (1.27) | (1.03) | (1.60) | (0.47) |
| | 8 | 67.23 | 76.48 | *72.61* | ***28.57*** | *39.13* | 36.81 |
| | | (2.68) | (3.36) | (1.42) | (0.92) | (1.51) | (0.51) |
| | 16 | 67.30 | 76.46 | 72.53 | 28.53 | 39.10 | *36.81* |
| | | (2.60) | (3.29) | (1.43) | (1.01) | (1.56) | (0.48) |

Among the three methods, RT-DETR achieves the best performance using 8 decoder heads (highlighted with boldface fonts in Table 9). The RT-DETR employs Hungarian matching instead of NMS, enabling improved performance at lower thresholds, as indicated by the number of decoder heads. The large difference in both AP and AR across different decoder heads suggests that RT-DETR is more sensitive to the number of heads, exhibiting a greater standard deviation. Deformable DETR exhibits similar behavior to RT-DETR due to its sparse attention. However, deformable DETR has the best balance between precision and recall, with higher F1 scores among the three methods and stability comparable to YOLOv11. YOLOv11 provides more precise and consistent detections with the same number of heads. A larger number of decoder heads results in little performance gain.

Overall, a smaller number of decoder heads, e.g., 8 or 4, benefits the detection of small objects. Many decoder heads may lead to fragmented feature representations that result in a drop in AR, indicating a decrease in valid detections. A balance between the number of sampling points and decoder heads helps avoid extraneous computation costs and performance drop.

### 4.5. Single- and Multi-scale Features

Table 11 reports the average AP, AR, F1, and standard deviations from methods that have single and multi-scale functionality through implementations of the pyramid network modules. In the case of using single-scale features, FPN modules are removed from the network. Using single-scale features, Faster R-CNN achieves the best performance with the least difference in F1 scores from multi-scale, which is attributed to its independent local feature refinement of each region by the RPN and ROI modules. However, comparing the performance of each method when single- and multi-scale features are used, implementing multi-scale features demonstrates an unarguable advantage. Using multi-scale and feature fusion by FPN enhances the detection of small objects, resulting in increased performance across all methods. Among all cases, YOLOv11 using multi-scale features (i.e., FPN) yields the best performance.

Table 11: AP, AR, and F1 using single and multi-scale.

| Scale | Method | 50 | | | 50:95 | | |
|---|---|---|---|---|---|---|---|
| | | AP | AR | F1 | AP | AR | F1 |
| Single scale | Faster R-CNN | **43.90** | **63.74** | **54.40** | **13.77** | **25.55** | **22.56** |
| | | (0.86) | (2.00) | (1.19) | (0.69) | (0.12) | (1.00) |
| | Cascade R-CNN | 6.10 | 23.86 | 17.64 | 1.23 | 6.03 | 4.50 |
| | | (0.24) | (1.69) | (0.31) | (0.05) | (0.62) | (0.12) |
| | YOLOv11 | 1.60 | 3.42 | 4.25 | 0.43 | 0.9 | 1.17 |
| | | (1.44) | (3.23) | (2.71) | (0.36) | (0.88) | (0.80) |
| | Def. DETR | 12.93 | 38.45 | 53.50 | 2.67 | 11.10 | 17.96 |
| | | (1.72) | (1.96) | (1.28) | (0.37) | (0.86) | (0.48) |
| | RT-DETR | 19.30 | 38.52 | 31.46 | 4.70 | 12.53 | 10.32 |
| | | (4.53) | (3.61) | (4.13) | (1.59) | (2.35) | (2.13) |
| Multi-scale | Faster R-CNN | 56.80 | 71.74 | 66.07 | 20.93 | 32.83 | 30.79 |
| | | (0.92) | (2.98) | (0.15) | (0.41) | (1.22) | (0.27) |
| | Cascade R-CNN | 47.93 | 65.16 | 41.20 | 17.97 | 29.57 | 15.76 |
| | | (2.12) | (2.52) | (1.44) | (0.86) | (1.19) | (0.73) |
| | YOLOv11 | **63.47** | **75.19** | 70.40 | **26.27** | **37.93** | 34.92 |
| | | (1.74) | (3.31) | (1.14) | (0.79) | (1.48) | (0.35) |
| | Def. DETR | 58.63 | 71.93 | **84.96** | 20.70 | 32.63 | **43.07** |
| | | (3.50) | (3.42) | (0.31) | (1.15) | (1.74) | (0.27) |
| | RT-DETR | 61.70 | 73.88 | 67.78 | 24.50 | 36.07 | 33.08 |
| | | (2.37) | (3.07) | (1.12) | (0.82) | (1.36) | (0.39) |

It is interesting to note that YOLOv11 and the DETR methods have at least a 20% boost in performance from using multi-scale feature fusion. While adding multi-scale and feature fusion techniques allows better detection of small objects, the amount of improvement varies, as shown in Table11.

### 4.6. Performance Comparison and Model Generalization

We conduct a quantitative and qualitative performance comparison of all six methods. Besides a performance comparison using xView data, our study includes method generalization via transfer learning using SkySat data. The number of backbone layers that are frozen is reported for each method when applying the trained weights for fine-tuning the SkySat dataset. Models are trained using xView examples; the same models are then applied to the validation sets of xView and SkySat for performance comparison and understanding the generalization ability. In addition, we report the computational costs in terms of the number of parameters, memory usage, and time. Each method is trained with its best combination of hyperparameters as shown in Table 12. SSD has its range of anchor box sizes set to 300, while the R-CNN methods have their range of anchor box sizes set to 90. The methods that use attention have the number of sampling points and decoder heads set to

8. The number of iterations of training across the methods is represented in the scale of $10^3$.

Table 12: Network hyperparameters. RABS: Range of Anchor Box Size.

| Method | RABS | Sampling points | Decoder heads | Itr. ($10^3$) | Frozen layers |
|---|---|---|---|---|---|
| SSD | 300 | - | - | 20 | 13 |
| Fas. R-CNN | 90 | - | - | 270 | 49 |
| Cas. R-CNN | 90 | - | | 270 | 49 |
| FocusDet | 326 | - | - | 59 | 10 |
| FFCA-YOLO | 326 | - | - | 29 | 10 |
| YOLOv11 | - | 8 | 8 | 94 | 10 |
| RT-DETR | - | 8 | 8 | 252 | 13 |
| Def. DETR | - | 8 | 8 | 252 | 49 |

Table 13 reports the average AP, AR, and F1 using the three datasets. The standard deviations are also reported in parentheses. In our experiments, models are trained from scratch using the xView dataset. For DOTA and SkySat datasets, fine-tuning the trained models (from xView) is performed due to the limited examples. The DOTA dataset has more examples than the SkySat dataset, which demonstrates an advantage in the refined models. The methods are in a descending order according to the AP and AR results of the xView dataset.

Among all six methods, YOLOv11 achieves the best performance, exhibiting the 9 best and 6 second-best scores out of 18 (AP, AR, and F1 combined). Its standard deviation is small and comparable to the other methods, which demonstrates consistent performance. RT-DETR yields a very competitive performance with 1 best and 7 second-best scores. In addition, Deformable DETR and Faster R-CNN produce good results in several cases. Overall, when there are a large number of training examples, e.g., the xView dataset, models with satisfactory performance can be created from YOLOv11 and transformer-based methods. The margin between the best and the second best is, although distinct, insignificant.

The results using DOTA and SkySat datasets demonstrate the transferability and generalization ability of the methods. In these experiments, models are trained using the xView dataset and fine-tuned with the examples of DOTA and SkySat. The overall performance of the methods using the DOTA dataset is comparable to the results of the models trained using the xView dataset. The fine-tuned YOLOv11 models achieved a better performance than the ones trained with the xView dataset, for example, $AP_{50}$ improves by 2.7% and $AP_{50:95}$ improves by 24.4%. The standard deviations are in a similar range to the models trained with the xView dataset. However, the evaluation results using the SkySat show a significantly lower performance. For example, $AP_{50}$ of YOLOv11 reduces to 14.75% and $AP_{50:95}$ reduces to 5.33%. This is attributed to the small dataset for model fine-tuning. As shown in Table 4, the SkySat dataset contains 55 training images, which is about one percent of the training size of the DOTA. FFCA-YOLO exhibits a bit better performance. Given that the target application of SkySat is detecting bee boxes, which is quite different from the application of the xView dataset, the generalization ability of all methods is limited. Additionally, the training data size clearly has a significant impact on the fine-tuned models.

Table 13: AP, AR, and F1 for small object detection. The backbone network of YOLOv11 and SSD is C3k2 and VGG-16, respectively. The rest methods use ResNet-50 as the backbone network. The number of parameters is in millions.

| Data | Method | 50 | | | 50:95 | | |
|---|---|---|---|---|---|---|---|
| | | AP | AR | F1 | AP | AR | F1 |
| | YOLOv11 | **69.47** | **78.60** | 73.97 | **29.58** | **40.30** | 37.63 |
| | | (3.07) | (3.47) | (1.81) | (1.40) | (1.78) | (1.04) |
| | RT-DETR | 68.20 | 78.31 | 72.97 | 28.15 | 39.22 | 36.48 |
| | | (3.06) | (3.44) | (1.85) | (0.73) | (1.77) | (1.04) |
| | Def. DETR | 62.20 | 74.35 | **85.79** | 23.43 | 36.45 | **46.09** |
| | | (2.72) | (2.93) | (0.20) | (1.05) | (1.55) | (0.27) |
| xView | Faster | 61.65 | 73.99 | 69.44 | 23.32 | 35.78 | 32.97 |
| | R-CNN | (3.22) | (3.22) | (2.07) | (1.30) | (1.77) | (1.09) |
| | Cascade | 55.60 | 70.31 | 63.22 | 21.53 | 33.93 | 30.71 |
| | R-CNN | (2.12) | (3.09) | (2.07) | (0.90) | (1.55) | (0.68) |
| | SSD | 13.15 | 27.61 | 25.89 | 3.35 | 9.07 | 9.16 |
| | | (1.37) | (2.66) | (2.62) | (0.30) | (0.99) | (0.94) |
| | FFCA-YOLO | 63.57 | 74.43 | 69.40 | 25.33 | 35.80 | 33.59 |
| | | (2.57) | (3.31) | (1.52) | (0.90) | (1.47) | (0.64) |
| | FocusDet | 11.37 | 34.16 | 22.80 | 8.50 | 11.20 | 7.28 |
| | | (4.04) | (7.12) | (5.50) | (0.87) | (2.62) | (1.91) |
| | YOLOv11 | **71.30** | **76.47** | 76.18 | **36.80** | **44.10** | 48.29 |
| | | (2.20) | (2.55) | (0.99) | (1.21) | (1.57) | (0.67) |
| | RT-DETR | 68.46 | 75.80 | 72.67 | 32.97 | 41.73 | 43.77 |
| | | (1.32) | (2.38) | (0.48) | (0.75) | (1.40) | (0.13) |
| | Def. DETR | 63.50 | 72.55 | **89.41** | 25.73 | 35.60 | **53.99** |
| | | (1.87) | (2.45) | (0.48) | (0.85) | (1.22) | (0.46) |
| DOTA | Faster | 65.00 | 73.62 | 69.66 | 29.73 | 38.73 | 41.59 |
| | R-CNN | (1.59) | (2.46) | (0.92) | (0.84) | (1.40) | (0.55) |
| | Cascade | 67.80 | 75.17 | 73.44 | 33.90 | 42.10 | 46.04 |
| | R-CNN | (2.06) | (2.79) | (1.55) | (0.92) | (1.43) | (0.78) |
| | SSD | 5.30 | 17.76 | 14.23 | 1.43 | 6.23 | 4.60 |
| | | (0.45) | (0.64) | (0.69) | (0.17) | (0.29) | (0.23) |
| | FFCA-YOLO | 38.43 | 54.67 | 45.60 | 15.57 | 26.47 | 21.00 |
| | | (0.58) | (1.31) | (0.87) | (0.24) | (0.87) | (0.41) |
| | FocusDet | 11.73 | 31.65 | 22.69 | 3.10 | 11.93 | 7.71 |
| | | (0.70) | (0.72) | (0.63) | (0.08) | (0.37) | (0.19) |
| | YOLOv11 | **14.75** | 38.63 | 23.91 | 5.33 | 14.30 | 8.89 |
| | | (10.91) | (19.01) | (9.87) | (4.83) | (8.33) | (4.75) |
| | RT-DETR | 11.78 | 24.67 | 18.01 | **5.53** | **25.48** | 9.92 |
| | | (7.81) | (35.82) | (10.18) | (3.50) | (19.51) | (6.33) |
| | Def. DETR | 4.65 | 16.67 | 11.70 | 1.46 | 6.60 | 4.66 |
| | | (4.67) | (14.84) | (11.25) | (1.57) | (5.07) | (4.37) |
| SkySat | Faster | 14.63 | 22.44 | **27.12** | 4.20 | 7.70 | 9.46 |
| | R-CNN | (7.78) | (14.02) | (11.93) | (1.61) | (4.00) | (3.07) |
| | Cascade | 6.67 | 16.97 | 15.76 | 1.73 | 5.57 | 5.14 |
| | R-CNN | (1.19) | (5.99) | (5.68) | (0.33) | (2.29) | (1.92) |
| | SSD | 0.43 | 6.32 | 3.07 | 0.18 | 2.07 | 1.06 |
| | | (0.45) | (9.30) | (2.64) | (0.21) | (3.02) | (0.97) |
| | FFCA-YOLO | 14.37 | **47.60** | 23.94 | 4.80 | 16.23 | **9.94** |
| | | (6.79) | (26.47) | (9.23) | (1.55) | (6.26) | (2.63) |
| | FocusDet | 0.15 | 2.23 | 1.42 | 0.03 | 0.52 | 0.41 |
| | | (1.23) | (1.17) | (0.62) | (0.01) | (0.06) | (0.12) |

Fig. 6 depicts six exemplary images of xView, SkySat, and DOTA with the ground truth marked with green boxes. The results are shown in Fig. 7, Fig. 8, and Fig. 9, in which the predictions are depicted with red boxes. Because detections are depicted with the ground truth (green boxes), in the case of an ideal detection, the detection (red boxes) completely occludes the ground truth. Any offset of detections and the ground truth indicates disagreements and localization errors.

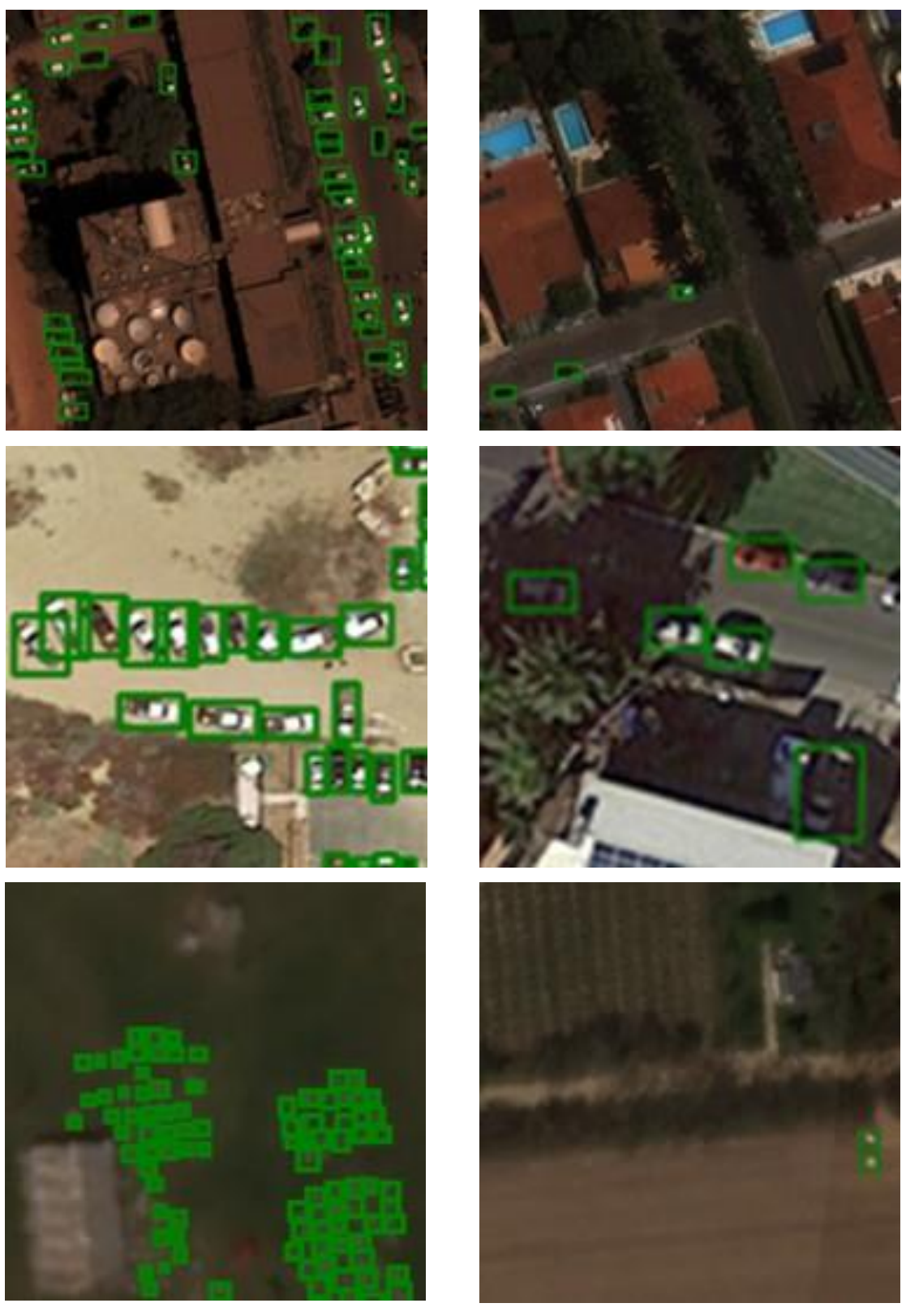

Figure 6: Exemplary images of xView (top), DOTA (middle), and SkySat (bottom).

In the two xView cases (as shown in Fig. 7), YOLOv11 exhibits superior performance in object detection and localization, which misses three objects in the top case and detects all in the bottom case. RT-DETR and Deformable DETR demonstrate excellent results with fewer missed detections. However, the detections face greater localization errors in the correct detections, as depicted by the misalignment of the red boxes (the detections) and the green boxes (the ground truth) underneath in the top case. False positives exist in the result of RT-DETR, as shown in the bottom case. The anchor-based methods, including SSD and R-CNN-based methods, face greater difficulty and generate many false positive detections and misses. SSD struggles the most to maintain alignment with its detections due to the usage of predefined anchor boxes, which is evident from the large number of misaligned detections.

Similar patterns can be observed in the results of DOTA and SkySat images, as shown in Fig. 8 and Fig. 9. However, as we have a much smaller number of examples of the SkySat images, we observe a significant number of missing detections, especially in the case of densely clustered small objects, as shown in the top cases of Fig. 9. Deformable DETR completely misses all objects in the top case. R-CNN-based methods appear to have more detections, but many of the detections are false positives. RT-DETR, on the other hand, fails to detect the small objects in the bottom case. Again, this signifies the importance of the volume of training examples, especially for the transformer-based methods. Fig. 10 depicts the zoom-in view of a failure case of each method. The cases for YOLOv11, FFCA-YOLO, Faster R-CNN, and FocusDet are dominated by missed detections. The cases for Cascade R-CNN and Drformable DETR are primarily false detections. The cases for SSD and RT-DETR are mixtures of false and missed detections.

Table 14: Computational efficiency.

| Method | Para. # (M) | VRAM (GB) Train | VRAM (GB) Infer | GPU (GFLOPs) Train | GPU (GFLOPs) Infer | Time (s) Train | Time (s) Infer |
|---|---|---|---|---|---|---|---|
| YOLOv11 | 25.37 | 4.15 | 0.69 | 607.66 | 55.73 | 56,104 | 44 |
| RT-DETR | 67.75 | 9.02 | 2.51 | 931.86 | 298.50 | 67,345 | 19 |
| Def. DETR | 40.95 | 3.59 | 1.56 | 785.69 | 324.11 | 46,500 | 59 |
| Fas. R-CNN | 41.29 | 10.31 | 0.54 | 387.70 | 245.11 | 590 | 36 |
| Cas. R-CNN | 69.09 | 18.99 | 0.90 | 441.51 | 262.26 | 1,507 | 83 |
| SSD | 24.18 | 16.77 | 0.28 | 525.29 | 175.08 | 2,261 | 235 |
| FFCA-YOLO | 7.13 | 2.72 | 1.01 | 31.60 | 15.80 | 9,211 | 13 |
| FocusDet | 2.33 | 16.70 | 2.96 | 14.70 | 7.40 | 8,796 | 9 |

Table 14 reports the model size and computational resources, including VRAM size, GPU floating-point operations, and time used by the six methods. The model size is represented by the number of parameters in millions, while the other three are represented by memory size, computation cost, and the time taken for training and inference per image. The memory is the peak GPU VRAM utilization of the method for processing a single $512 \times 512$ image with a batch size of 1 for comparison. The training and inference times are the total time taken to train the model in seconds. Our experiments are conducted on a system with an RTX 4090 GPU with 24GB of VRAM, an Intel i9 CPU, 64GB of memory, and Ubuntu. The compared methods are programmed with Python 3.8, PyTorch 2.4.2, and CUDA 11.8.

Among the methods compared, YOLOv11 and SSD stand out for their relatively small parameter sizes of around 25 million, making them particularly attractive for edge deployments such as satellites, drones, or ground stations where onboard memory is limited. In contrast, RT-DETR and Cascade R-CNN require parameter counts in the upper 60 million range due to their additional decoder layers and detection heads, which place a heavier demand on storage and computational resources. Parameter size is a critical factor in deployment, as smaller models not only reduce storage requirements but also allow for more efficient execution on devices with restricted resources.

While SSD benefits from its compact size, it suffers from a high training memory footprint, which makes on-device training less practical. Nevertheless, its inference memory footprint is relatively small, which makes SSD useful for holding pre-trained models tailored to detecting small objects of consistent shapes. In contrast, two-stage detectors such as Faster R-CNN and Cascade R-CNN require substantial memory during training because of the large number of proposals generated by the RPN. Transformer-based approaches, e.g., RT-DETR and Deformable DETR, shift the burden to inference, consuming more memory because of cross- and self-attention mechanisms. YOLOv11 strikes a middle ground, requiring only 4.15 GB for training and 0.69 GB for inference, while Deformable DETR asks for a smaller training memory usage at 3.59 GB. These characteristics make YOLOv11 and Deformable DETR partic-ularly suitable for edge devices that must handle both training and inference within strict memory limits.

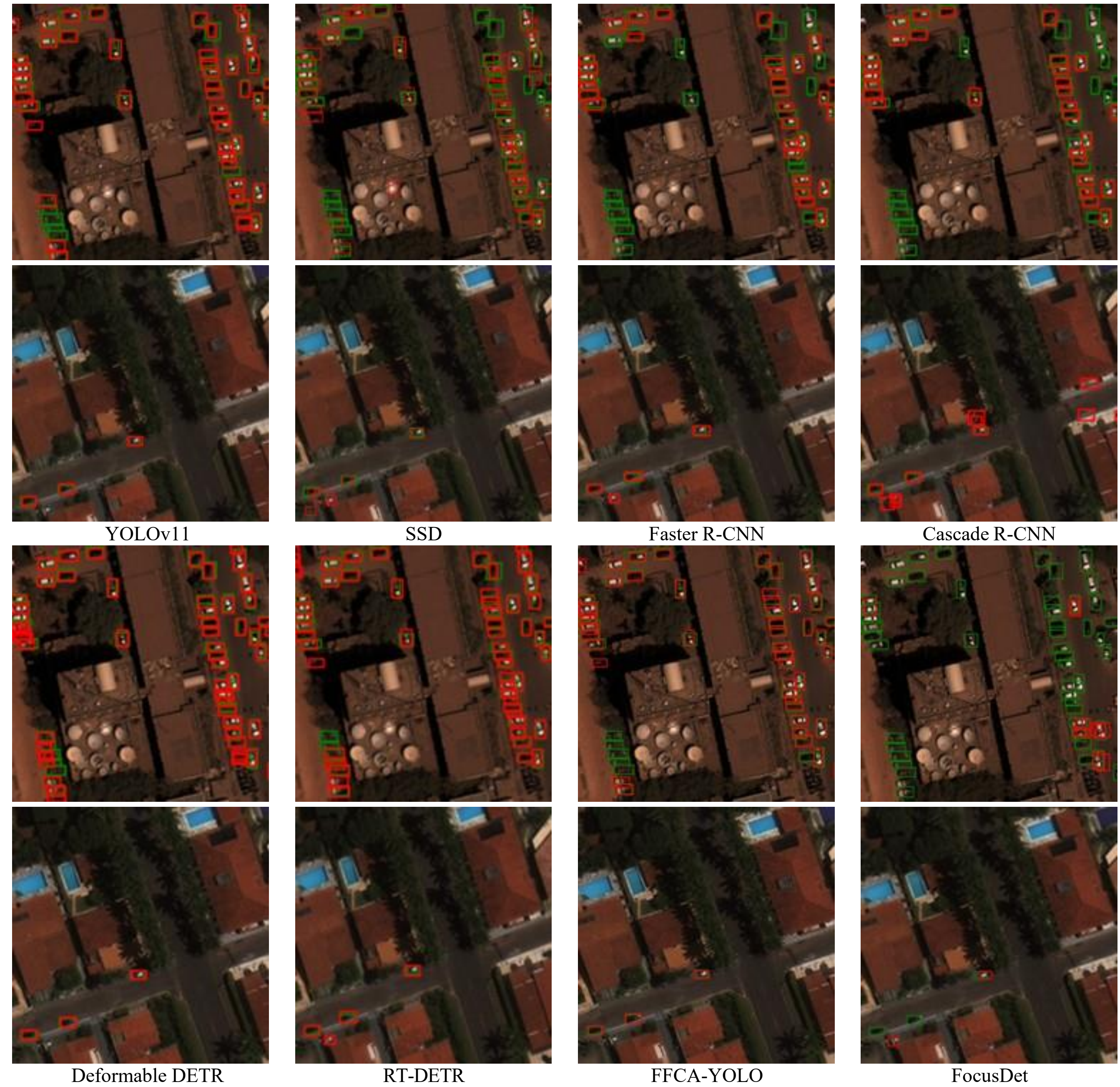

Figure 7: Results of small object detection on xView. Green and red boxes represent ground truth and detections, respectively.

The computational cost also plays a decisive role in determining deployment strategies. Transformer-based models are the most expensive in terms of GPU operations, with RT-DETR and Deformable DETR requiring 931.86 and 785.69 GFLOPs, respectively, during training. Their inference costs are similarly high, at 298.50 and 324.11 GFLOPs, compared to convolutional models. By contrast, R-CNN and SSD methods achieve faster training despite their larger memory footprints. Faster R-CNN, for instance, completes training in 590 seconds, followed by Cascade R-CNN at 1,507 seconds. Inference times, however, converge more closely across methods. RT-DETR delivers the fastest inference at 19 seconds, while Faster R-CNN follows at 36 seconds. YOLOv11, though slower to train, is optimized for inference with a relatively fast 44-second runtime. Cascade R-CNN incurs additional delays due to its complex detection heads, while SSD is hindered by its reliance on extensive pre- and post-processing.

These trade-offs highlight distinctions between real-time and offline use. For real-time deployment, such as onboard processing in satellites or drones, fast inference and efficient memory use are paramount. RT-DETR is particularly well-suited in this setting, combining rapid inference with high detection accuracy. YOLOv11 provides a balanced alternative, especially when modest memory and reliable inference are required with-out the need for frequent fine-tuning. SSD may also be viable in real-time scenarios when a pre-trained model is sufficient, but its limitations in on-device retraining make it less flexible. On the other hand, for offline or batch analysis conducted at ground stations, resource constraints are less restrictive, allowing heavier models to be used. Cascade R-CNN and Faster R-CNN become more viable in such settings, as their longer training times and higher memory requirements can be absorbed by high-performance hardware. Similarly, transformer-based models like RT-DETR and Deformable DETR excel in offline analysis, where their computationally intensive architectures are supported by dedicated GPUs, and their fast inference speeds help accelerate large-scale data processing.

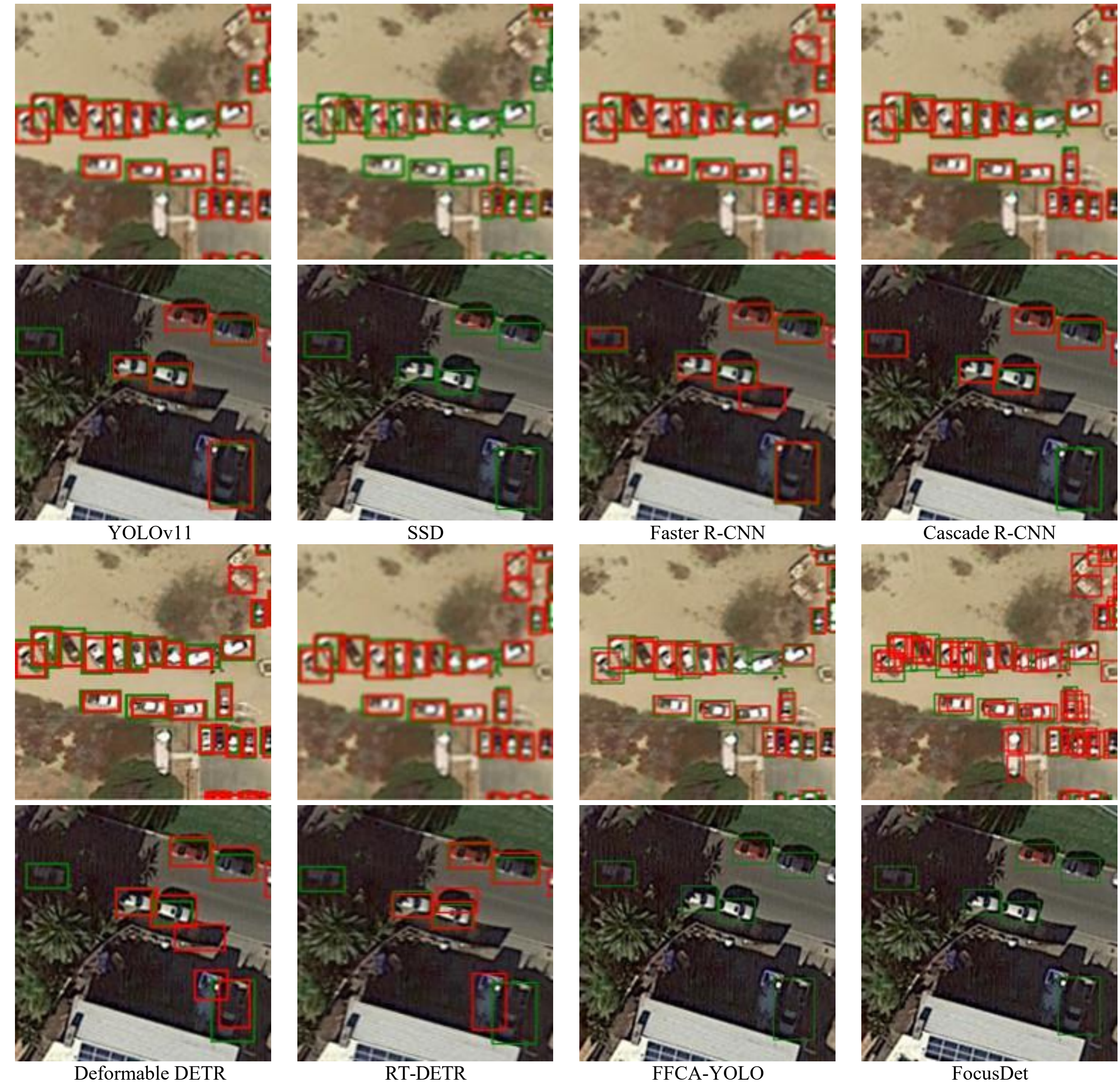

Figure 8: Results of small object detection on DOTA. Green and red boxes represent ground truth and detections, respectively.

### 4.7. Discussion

Anchor-based detectors can identify small objects when anchor sizes and aspect ratios are finely tuned, but this dependence limits their flexibility. Misaligned anchors lead to false positives and missed detections, especially in cluttered scenes, and stride constraints in models like SSD further trade off spatial detail for efficiency. Because these methods rely on dataset-specific anchor priors and strong gradient responses, they show limited generalizability across domains.

R-CNN methods improve localization by refining region proposals before classification, and their adaptive RoI pooling generally enhances generalization. However, their performance drops in scenes with many small, densely clustered objects.

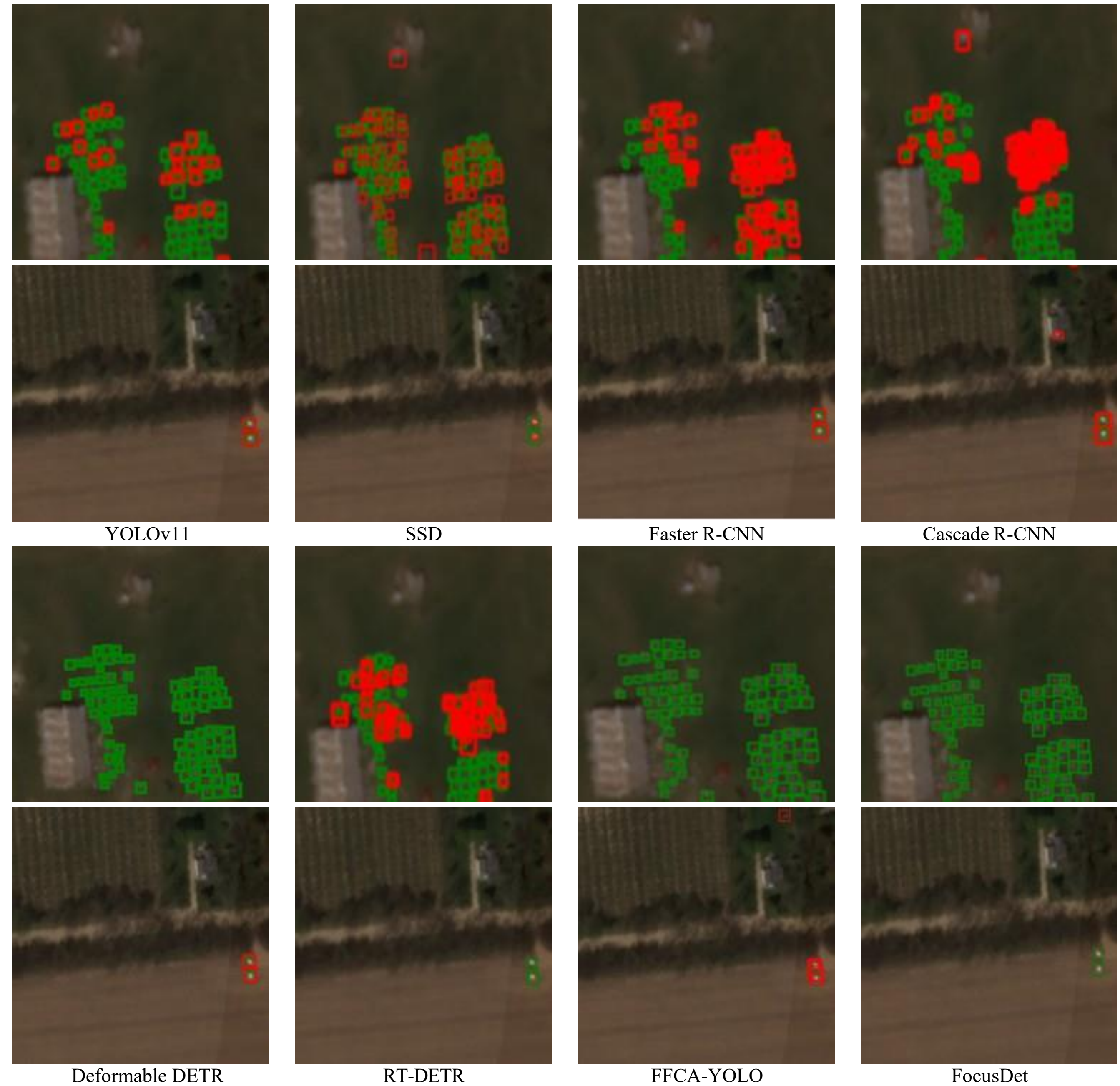

Figure 9: Results of small object detection on SkySat. Green and red boxes represent ground truth and detections, respectively.

Fixed-size RoI pooling distorts tiny features, suppressing true positives through NMS, as seen in both high-resolution and lower-resolution datasets. Cascade R-CNN aggravates this sensitivity because each stage depends on the quality of the previous one; poor RPN proposals for small objects propagate errors, causing either excessive or missed detections.

Anchor-free methods avoid handcrafted anchors by predicting boxes from center points or queries. YOLOv11 improves small-object alignment through task-aligned and distribution focal loss but remains sensitive to domain shifts. DETR-based models eliminate anchors using set-based prediction and multiscale attention. Deformable DETR improves generalization via sparse, multi-level sampling but can mishandle adjacent objects, leading to missed detections during Hungarian matching. RT-DETR addresses these issues with adaptive query refinement, though reducing the query for efficiency increases vulnerability to noisy gradients in cluttered backgrounds.

Occlusion and shadows challenge detections by introducing misleading edges. Attention-based models mitigate this by leveraging context. YOLOv11's localized attention helps isolate unobstructed regions but struggles without long-range dependencies. Deformable DETR's sparse sampling recovers visible cues but often overlooks occluded objects in dense scenes. RT-DETR is more robust due to its adaptive queries, which better separate partially visible objects from background textures.

Post-processing plays a central role in many methods. Non-maximum suppression in anchor-based and hybrid detectors removes redundant boxes but can inadvertently eliminate true detections in crowded regions. Although DETR-based models avoid it through Hungarian matching, matching becomes costly and unstable when objects lie extremely close to one another, resulting in miss-detections in dense clusters.

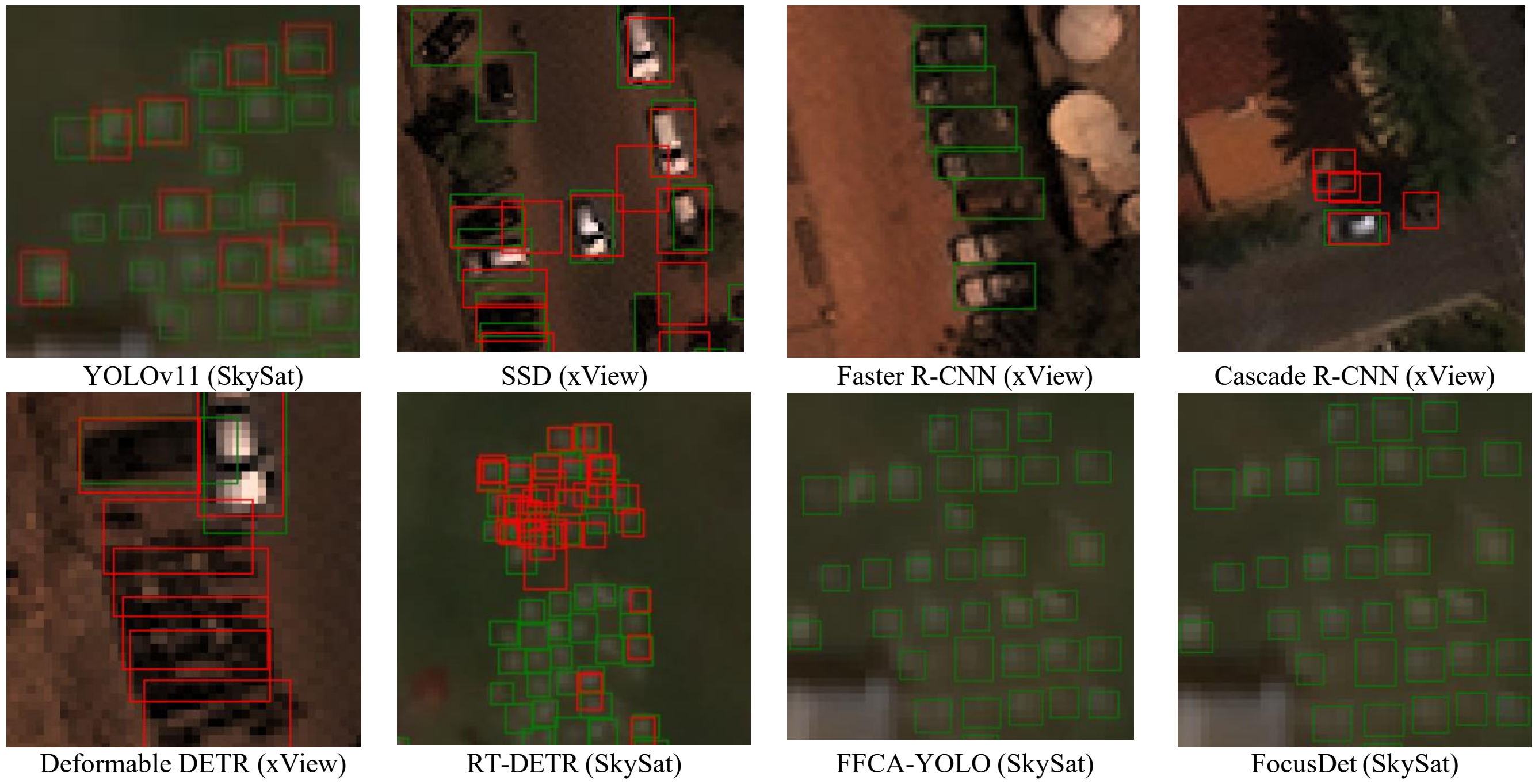

Figure 10: Zoom-in view of a failure case of each method. Green and red boxes represent ground truth and detections, respectively.

In summary, the employment of multiscale attention, context integration, and anchor-free prediction (e.g., variants of YOLO and DETR) offers a balance of robustness, generalization, and small-object sensitivity across diverse remote-sensing imagery. For applications with variable object scales or domain shifts, anchor-free and transformer-based detectors are preferable because the mismatched anchors often cause performance drops. When dealing with dense clusters of small objects, methods with high spatial fidelity, such as anchor-free designs, tend to outperform heavy cascades that amplify proposal errors. For imagery affected by occlusion or shadows, contextual or adaptive-query reasoning recovers signals from partially visible objects, which makes such methods (e.g., deformable and RT-DETR variants) better suited. In addition, post-processing is an important design choice. Tuning NMS thresholds or stabilizing Hungarian matching helps avoid misdetections, especially in dense clusters.

## 5. Conclusion

This paper presents an empirical evaluation of deep learning methods for small object detection from satellite imagery. Small objects are represented with a few pixels, and their detection faces great challenges due to the lack of descriptive features. Drawing from existing surveys and literature, we identified top-performing methods and conducted a thorough empirical evaluation using xView, DOTA, and SkySat images. In addition to a comparison across all methods using two different datasets, we investigate the impact of the choice of parameters and key components, including anchor box size, region proposal thresholds, attention mechanism, and feature fusion. We also compare the computational efficiency across models.

In our evaluation, YOLOv11 demonstrates a balance between performance and localization. Anchor-based models such as SSD, Faster R-CNN, and Cascade R-CNN remain highly sensitive to anchor box size, often requiring extensive fine-tuning to adapt to different object scales. While the inclusion of attention mechanisms and multi-scale feature representations enhances detection accuracy, these improvements come with diminishing returns relative to the computational cost. Transformer-based models, particularly RT-DETR and Deformable DETR, highlight both the promise and limitations of attention-driven designs. RT-DETR performs robustly under conditions of partial obstruction from vegetation or shadows, whereas Deformable DETR struggles with dense clusters of small objects and shows a higher susceptibility to dataset-specific overfitting. These findings point to a trade-off: anchor-based approaches demand careful design and tuning of prior assumptions, while transformer-based methods require substantially greater computational resources. Memory efficiency also remains uneven across architectures: SSD and R-CNN variants exhibit heavier memory footprints during training despite shorter runtimes, whereas attention-based models incur higher computational costs but deliver faster inference, with RT-DETR and YOLOv11 achieving inference times of 19s and 44s, respectively, close to Faster R-CNN's 36s.

The development of small object detection methods is moving toward hybrid architectures that integrate the strengths of convolutional backbones with attention mechanisms. YOLOv11 exemplifies this by incorporating attention into an efficient single-stage framework, while RT-DETR demonstrates how transformers can be optimized for faster inference without prohibitive computational costs. A second trend lies in improving efficiency through model compression, pruning, and quantization, which will be crucial for edge deployments on satellites and drones where resources are severely constrained. Additionally, emerging research is exploring task-specific optimizations, such as context-aware modules for dense object scenes and adaptive feature selection mechanisms for small-scale structures. Finally, the field is shifting toward multi-modal and cross-domain learning, where detectors can leverage complementary sources such as multispectral imagery.

In addition, the development of small object detection is likely to be shaped by advances in foundation models and large-scale pretraining. Current detectors are still limited by their reliance on task-specific architecture and labeled datasets, which can constrain generalization across domains. Foundation models trained on vast amounts of multimodal data, including images, text, and video, offer a pathway to richer feature representations that may better capture the subtle cues of small objects. Self-supervised learning approaches are also gaining traction, enabling models to learn from large pools of unlabeled data and reducing dependence on costly annotations. This trend could be particularly transformative for small object detection, where annotated examples are often scarce and expensive.

Ultimately, small object detection is moving toward a paradigm that blends efficiency, generalization, and multimodal reasoning. The convergence of foundation models, hybrid architectures, and cross-domain learning will likely define the next generation of detectors, making them more robust to environmental variability and across platforms ranging from satellites and drones to large-scale ground systems.